\documentclass{ecai}  

\usepackage{graphicx}
\usepackage{latexsym}
\usepackage{amsmath,amsfonts}
\usepackage{array}
\usepackage{url}
\usepackage{verbatim}
\usepackage{hyperref}
\usepackage{academicons}
\usepackage{amsmath}
\usepackage{xcolor}
\usepackage{longtable}
\usepackage{caption}
\usepackage{multirow}
\usepackage{adjustbox}
\usepackage{rotating}
\usepackage{neuralnetwork}
\usepackage{tabularx}
\usepackage{lipsum}
\usepackage{algorithm}
\usepackage{algpseudocode}

\usepackage{rotating}
\usepackage{pgfplots}
\usepackage{graphicx,subcaption}

\newcommand{\BibTeX}{B\kern-.05em{\sc i\kern-.025em b}\kern-.08em\TeX}

\begin{document}


\begin{frontmatter}


\paperid{123} 



\title{Scaling Transformers for Time Series Forecasting: Do Pretrained Large Models Outperform Small-Scale Alternatives?}


\author[A]{\fnms{Sanjay}~\snm{Chakraborty}\thanks{Corresponding Author. Email: sanjay.chakraborty@liu.se.}}
\author[B]{\fnms{Ibrahim}~\snm{Delibasoglu}}
\author[C]{\fnms{Frerik}~\snm{Heintz}} 

\address[ABC]{Department of Computer and Information Science (IDA), REAL, AIICS, Linköping University, Sweden}

\begin{abstract}

Large pre-trained models have demonstrated remarkable capabilities across domains, but their effectiveness in time series forecasting remains understudied. This work empirically examines whether pre-trained large-scale time series models (LSTSMs) trained on diverse datasets can outperform traditional non-pretrained small-scale transformers in forecasting tasks. We analyze state-of-the-art (SOTA) pre-trained universal time series models (e.g., Moirai, TimeGPT) alongside conventional transformers, evaluating accuracy, computational efficiency, and interpretability across multiple benchmarks. Our findings reveal the strengths and limitations of pre-trained LSTSMs, providing insights into their suitability for time series tasks compared to task-specific small-scale architectures. The results highlight scenarios where pretraining offers advantages and where simpler models remain competitive.

\end{abstract}

\end{frontmatter}

\section{Introduction}
Time series analysis is a crucial technique in data science, enabling insights into temporal data patterns. It encompasses forecasting, which predicts future values based on historical trends, aiding applications like stock market prediction \cite{zhang2024hybrid}, land-use monitoring \cite{shi2025digital}, energy consumption \cite{yaprakdal2023multivariate}, and weather forecasting \cite{engel2024transformer}. Time series can be classified as either univariate or multivariate, describing one or more variables that change over time, respectively \cite{zhang2024self}. Time series forecasting is further categorized into short-term and long-term forecasting based on the prediction horizon. Short-term forecasting predicts values over a brief period (e.g., minutes to days) and is used in real-time applications like stock trading and energy demand prediction. In contrast, long-term forecasting extends over weeks to years, capturing broader trends for applications such as climate modeling and economic forecasting \cite{li2024functional,li2024deep}. Both require effective modeling of temporal dependencies to ensure accurate predictions. The scene has changed dramatically since deep learning emerged, bringing with it the incorporation of deep neural networks into time series challenges. These strategies have outperformed conventional forecasting techniques by making use of the greater availability of data and computing power. Transformers \cite{vaswani2017attention} stand out among them, encouraging innovation in a wide range of fields \cite{ahmed2023transformers}. They can do so with ease thanks to their unique stacking structure and self-attention processes \cite{wen2022transformers}. Consequently, transformers demonstrate superior performance not only in forecasting tasks but also in representation learning, where they extract abstract features that can be leveraged for downstream tasks like anomaly detection and classification \cite{yang2024breaking}. 
Large-scale Time Series Models (LSTSMs) have surpassed traditional transformers in various tasks due to their scalability, extensive pretraining, security, and adaptive capabilities \cite{abdel2024federated}. Unlike standard transformers, which process fixed-length sequences with self-attention, large-scale models leverage massive datasets, increased parameter counts, and architectural innovations such as sparse attention and retrieval-augmented generation. This enables them to generalize better, capture long-range dependencies, and adapt to diverse downstream applications, including reasoning, anomaly detection, and classification. While large-scale models require significant computational resources, their superior performance in complex tasks makes them a powerful alternative to conventional transformer-based models \cite{ahamed2024timemachine}. Zero-shot, one-shot, few-shot, and full-shot learning define how large-scale models generalize to new tasks based on prior knowledge and minimal examples. In zero-shot learning, the model performs a task without any prior examples, relying solely on its pre-trained knowledge. One-shot learning provides a single example before prediction, while few-shot learning uses a handful (typically 2–10) of examples for better adaptation. Full-shot learning (fine-tuning) involves training the model on a large dataset for a specific task, ensuring high accuracy. While pretraining builds a general-purpose foundation, fine-tuning adapts the model to specialized domains or tasks. The key differences between large-scale models and conventional transformers are discussed below,

\textit{A. Architecture Differences:}  
Transformers and Large-scale Time Series Models (LSTSMs) are both based on self-attention mechanisms, but their use in time series tasks can differ due to the data's structure and the nature of the tasks. In Transformers, the attention mechanism processes all time steps simultaneously, allowing the model to capture both past and future information \cite{liu2025autotimes}. In contrast, large-scale models inspired by large language models (LLMs), like GPT, process the time series data in an autoregressive manner, predicting future time steps based on past observations only. Mathematically, for Transformers, the attention scores are computed as:
\[
A = \text{softmax}\left( \frac{QK^T}{\sqrt{d_k}} \right)
\]
whereas large-scale models predict future values in a causal manner as:
$\hat{x}_{T+1} = f_{\text{LSTSM}}(x_1, x_2, \dots, x_T)$
showing the distinction between bidirectional and unidirectional processing of data.

\textit{B. Handling Sequential Data:}  
Transformers typically attend to all time steps in a sequence simultaneously, processing data from both the past and future. This makes them suitable for tasks like time series classification, where context from the entire sequence is important. On the other hand, large-scale models operate autoregressively, utilizing only past time steps to predict future values. This is particularly useful for tasks like time series forecasting, where only past data is relevant \cite{woo2024unified,ansari2024chronos}. 

\textit{C. Task-Specific Adaptations:}  
For time series classification tasks, Transformers process the entire sequence to learn temporal patterns, yielding a global representation for decision-making. The classification layer is typically a softmax function:
\[
\hat{y} = \text{softmax}(W_C \cdot \text{Encoder}(X))
\]
whereas large-scale models, primarily focused on autoregressive tasks, learn to predict future time steps from past observations, making them especially effective for forecasting and anomaly detection. In anomaly detection, large-scale models can flag anomalous values based on prediction errors by comparing predicted values to actual ones, while Transformers rely on their attention mechanism to identify outliers within the sequence \cite{goswami2024moment}.

\textit{D. Model Efficiency and Complexity:}  
Transformers are highly expressive but computationally expensive due to the quadratic complexity of self-attention. For long time series, the attention matrix grows as \( O(T^2) \), which becomes computationally prohibitive for very large sequences. LSTSMs, however, are more efficient for autoregressive tasks, as they only attend to past time steps for prediction, though they still exhibit inefficiencies in predicting multiple time steps. The computational cost for Transformers scales with the entire sequence, whereas LSTSMs scale with the autoregressive nature of prediction, processing one step at a time \cite{liu2024timer,liu2024timerxl}.

\textit{E. Key Differences in Time Series Tasks:}  
For time series forecasting, Transformers are suitable for multi-step forecasting tasks, where the model can predict multiple future time steps simultaneously using bidirectional attention. Large-scale models, by contrast, excel in one-step-ahead forecasting tasks, predicting the next value based on the previous ones in a unidirectional manner. For time series classification, Transformers perform better by processing the entire sequence at once, capturing long-range dependencies to make classification decisions. LSTSMs, while capable of handling classification tasks, are less commonly used for this purpose and typically focus on sequence-generation tasks \cite{jin2023time,liang2024foundation,shi2024time}. 

\begin{table*}[hbt!]
\scriptsize
\caption{Comparative Analysis of LSTSMs and small-scale Transformers for Time Series Analysis.}
\begin{center}
\begin{tabular}{|c|c|c|}
\hline
\textbf{Feature}                       & \textbf{Large-scale models (LSTSMs)} & \textbf{Small-scale Transformers} \\ \hline
Pre-training on Large Data\cite{chang2023llm4ts}     & \checkmark (Leverages vast general knowledge)                 & $\times$ (Typically trained from scratch)            \\ \hline
Zero-shot/Few-shot Learning\cite{woo2024unified}   & \checkmark (Can generalize without retraining)                & $\times$ (Needs task-specific fine-tuning)           \\ \hline
Long Sequence Modeling \cite{liu2024timerxl}        & \checkmark (Expanding context window)                         & \checkmark (Positional encoding helps)                 \\ \hline
Computational Efficiency      & $\times$ (Very resource-intensive)                          & \checkmark (More optimized for time series)            \\ \hline
Interpretability (XAI)\cite{goswami2024moment}        & $\times$ (Black-box, hard to explain)                       & \checkmark (Attention mechanisms, feature attribution) \\ \hline
Handling Missing Data         & \checkmark (Can infer missing points contextually)            & \checkmark (Models gaps explicitly)                    \\ \hline
Multivariate Time Series\cite{li2024functional,liu2024calf}      & $\times$ (Struggles with multiple independent variables)    & \checkmark (Explicitly handles multiple features)      \\ \hline
Real-time Inference           & $\times$ (Slow, high latency)                               & \checkmark (Optimized for real-time applications)      \\ \hline
Autoregressive Forecasting\cite{liu2025autotimes}    & $\times$ (Not inherently designed for forecasting)          & \checkmark (Well-suited for forecasting tasks)         \\ \hline
Scalability to Large Datasets\cite{abdel2024federated} & $\times$ (Memory-intensive, expensive)                      & \checkmark (More efficient)                            \\ \hline
Generalization Across Domains [1ite{liu2024calf} & \checkmark (Pre-trained on diverse datasets)                   & $\times$ (Needs fine-tuning for new domains)         \\ \hline
Anomaly Detection\cite{ansari2024chronos}             & \checkmark (Context-aware anomaly detection)                  & \checkmark (Captures deviations in patterns)           \\ \hline
Training Cost\cite{das2024decoder}                 & $\times$ (Very high for fine-tuning)                        & \checkmark (Relatively lower)                          \\ \hline
Classification\cite{ansari2024chronos}                & \checkmark                                                    & \checkmark                                             \\ \hline
Federated Learning\cite{abdel2024federated}                & \checkmark (data privacy, communication overhead)                                                    & $\times$ (lack of data privacy)                                             \\ \hline
\end{tabular}
\label{comptab}
\end{center}
\end{table*}

\begin{figure}[hbt!]
\begin{center}
\includegraphics[scale=0.11]{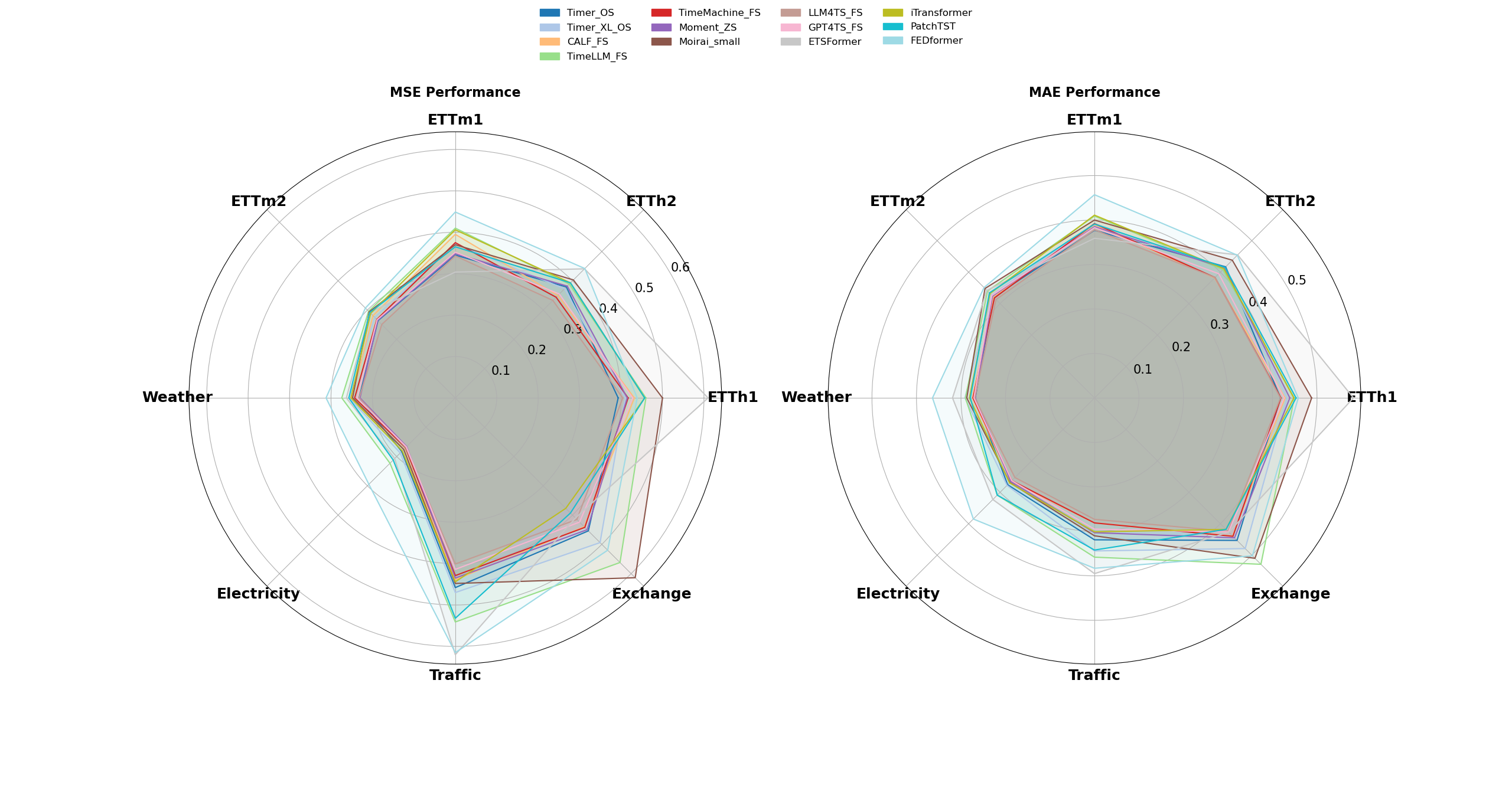}
\caption{Performance on long-term forecasting by SOTA LSTSMs (training from scratch and Small-scale Transformers)}
\label{Figradarlong1}
\end{center}
\end{figure}

\begin{figure}[hbt!]
\begin{center}
\includegraphics[scale=0.10]{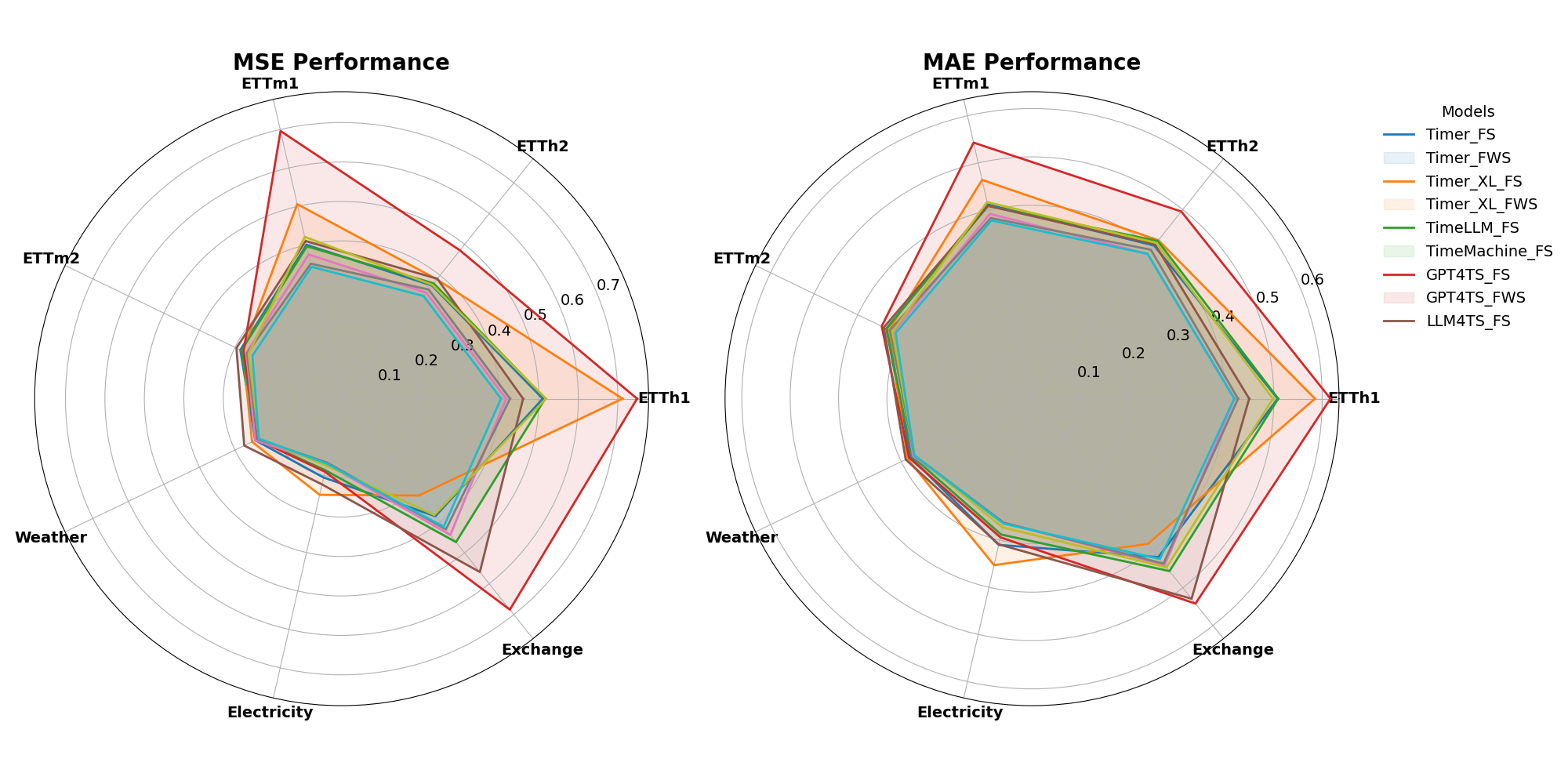}
\caption{Performance on long-term forecasting by SOTA LSTSMs (pretraining and fine-tuning) and Small-scale Transformers}
\label{Figradarlong2}
\end{center}
\end{figure}

\begin{figure}[hbt!]
\begin{center}
\includegraphics[scale=0.11]{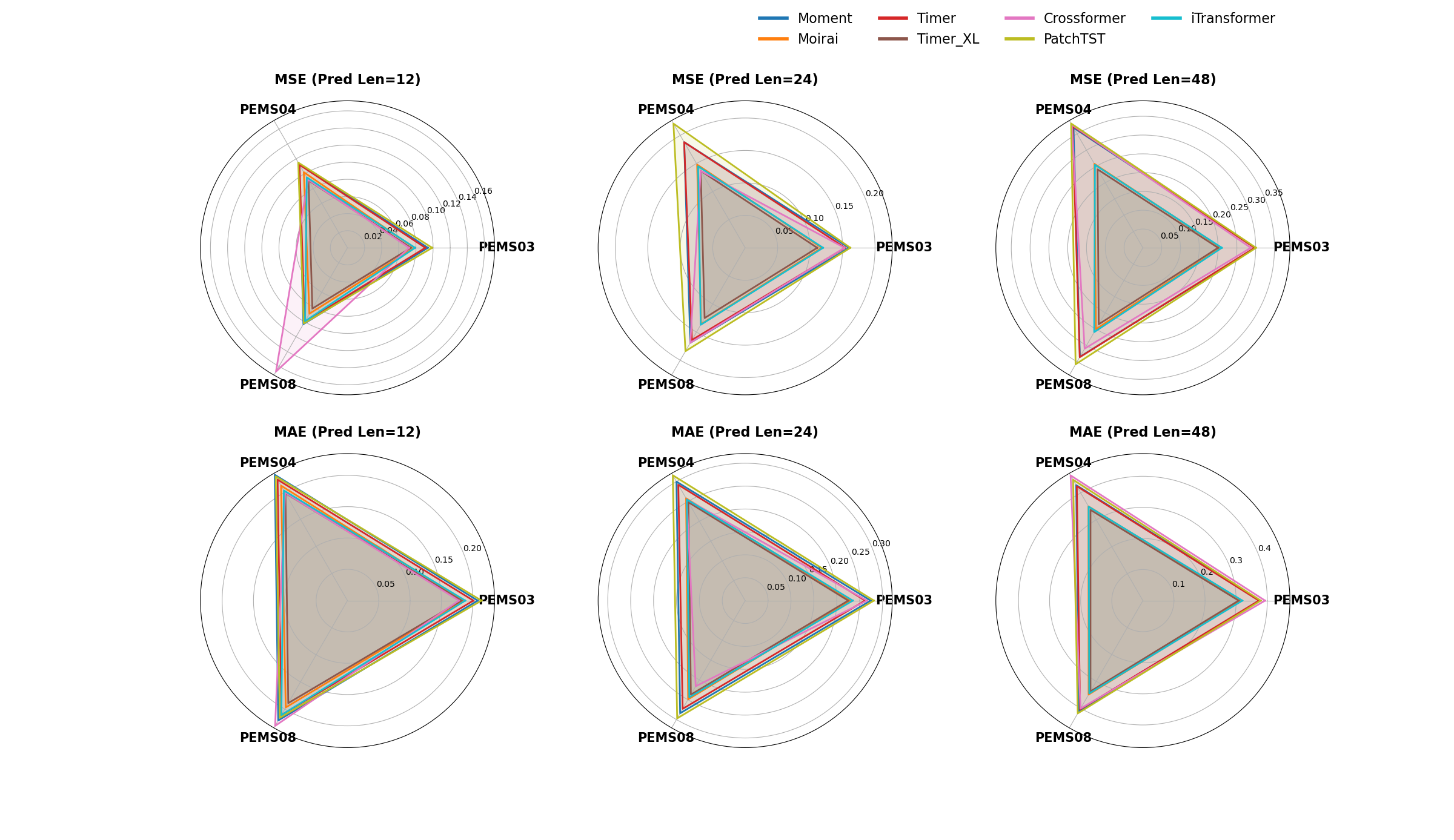}
\caption{Performance on short-term forecasting by SOTA LSTSMs and Small-scale Transformers}
\label{Figradarshort}
\end{center}
\end{figure}

Inspired by the above discussion, we have explored and analyzed a range of recently introduced pre-trained large-scale frameworks for time series forecasting, conducting an extensive empirical evaluation on several widely used time series datasets. The primary objective of this paper is to enhance readers' understanding of the differences between pre-trained large-scale time series models (LSTSMs) and small-scale transformers in time series forecasting. Specifically, the paper explores how these two approaches compare in terms of performance, including accuracy, efficiency, model structures, training behavior, and generalization across various (long-term and short-term) time series forecasting tasks. Additionally, it examines their applicability to different types of time series data, assessing whether LSTSMs provide advantages over training transformers from scratch or fine-tuning them on task-specific data. 
Through this comparison, the paper aims to clarify the strengths, limitations, and practical considerations of using these models in real-world time series applications. Figure \ref{Figradarlong1}, Figure \ref{Figradarlong2}, and Figure \ref{Figradarshort} demonstrate a performance comparison of the experimented state-of-the-art (SOTA) large-scale and small-scale language models across various benchmark datasets for long-term and short-term forecasting tasks, respectively. In summary, the paper’s main contributions are as follows:
\begin{itemize}
    \item We have examined the strengths and weaknesses of LSTSMs in comparison to small-scale time series transformers, highlighting key opportunities where LSTSMs can be valuable and impactful.
    \item We have explored recent advancements in large-scale models relevant to time series forecasting, with a primary focus on multivariate time series forecasting. Additionally, we have conducted extensive experiments on various benchmark datasets to provide empirical evidence on whether pre-trained LSTSMs and no-pretrained transformers are better suited for time series forecasting tasks. We have analyzed and compared both long-term and short-term time series forecasting for a set of benchmark datasets in this work.
    \item To the best of our knowledge, this is a complete study to analyze and compare a set of state-of-the-art (SOTA) Long-Term Sequence Learning Models (LSTSMs) with small-scale transformers for time series forecasting tasks, focusing on computational complexity, architectural differences, and performance quality.
\end{itemize}

\section{Background}
Time series can be classified as either univariate or multivariate \cite{montgomery2015introduction}. A univariate time series consists of observations of a single variable recorded over time, such as daily temperature or stock prices, and is analyzed using models like ARIMA or simple RNNs. In contrast, a multivariate time series involves multiple interdependent variables evolving together, such as weather data (temperature, humidity, wind speed) or financial indicators (stock prices, interest rates), where relationships between variables improve forecasting accuracy. While univariate models focus solely on past values of one variable, multivariate models, like VAR, LSTMs, and Transformers, capture cross-variable dependencies, making them more powerful for complex real-world applications \cite{zhang2024self}.

Time series forecasting can be classified as either long-term or short-term. Long-term forecasting involves predicting trends, patterns, or values over an extended period, often months or years and is commonly used in climate modeling, economic planning, and energy demand estimation. It focuses on capturing broader trends and structural patterns but may suffer from accumulating errors and uncertainty \cite{liu2021forecast}. In contrast, short-term forecasting predicts values over a much shorter horizon, typically ranging from minutes to weeks and is widely applied in stock market predictions, real-time monitoring, and supply chain management. Short-term forecasts prioritize high-frequency patterns and recent trends, often achieving greater accuracy but with limited long-term generalizability. Combining both approaches can enhance decision-making by balancing immediate responsiveness with strategic foresight \cite{montgomery2015introduction}.

\section{Related Work}
Table \ref{comptab} highlights the strengths and weaknesses of 'Large-scale Time Series Models (LSTSMs)' and Transformers across various key features, including computational efficiency, generalization, forecasting capabilities, and interoperability. 
\subsection{Small-scale Time Series Transformers}
Transformer-based techniques for time series forecasting have been the center of a significant amount of study \cite{kitaev2020reformer, liu2022non, zeng2023transformers, zerveas2021transformer, zhang2024multivariate}.  ETSformer \cite{woo2022etsformer}, in contrast to ordinary transformers, uses two distinct attention mechanisms—Frequency Attention (FA) for capturing periodic patterns and 'Exponential Smoothing Attention (ESA)' for long-term dependencies—to successfully describe trend and seasonality components. A dimension-segment-wise (DSW) method for encoding time-series data into a structured 2D representation is presented by Crossformer \cite{zhang2023crossformer}. The Sub-Window Tokenizer is used.  By using feed-forward networks to learn nonlinear representations and encoding individual time points as variate tokens, iTransformer \cite{liu2023itransformer} innovates and allows the attention mechanism to describe multivariate correlations. In order to improve the extraction of long-term information from the data, FEDformer \cite{zhou2022fedformer} uses the fast Fourier transform to break down sequences. Time series data are divided into subseries-level patches via PatchTST \cite{nie2022time}, which act as input tokens for a transformer-based architecture. In order to process many time series efficiently, it also uses channel independence, in which each channel contains a single univariate time series that has the same Transformer weights and embedding as all other series. An interesting work has been done on multivariate time series classification by integrating conformal predictors using possibilistic reasoning \cite{campagner2024ensemble}. Traditional transformers rely on fixed or learned positional embeddings, which may not effectively represent the sequential nature of time series. A work explores enhancements in position encoding techniques to better capture temporal dependencies in multivariate time series data \cite{foumani2024improving}. Another research work presents a hybrid model that integrates convolutional layers with transformers to enhance feature extraction and classification accuracy. This approach leverages sliced and channel-wise processing \cite{wu2022aggregated}. 
\subsection{Large-scale Time Series Models (LSTSMs)}
Recently, several pre-trained Large-scale Time Series Models (LSTSMs), including the generative pre-trained Transformer (GPT) of different versions \cite{zhou2023one,liu2024timer} and LLaMA \cite{chang2023llm4ts,liu2024calf} models, which are based on the transformer architecture, have emerged as powerful tools for generating high-quality outputs in time series analysis. 
TIME-LLM \cite{jin2023time} repurposes pre-trained Large Language Models (LLMs) for time series forecasting by transforming temporal data into a format compatible with language models. This approach enables LLMs to capture complex temporal dependencies efficiently, leveraging their deep contextual understanding without requiring task-specific training from scratch. AutoTimes \cite{liu2025autotimes} introduces a framework that leverages Large Language Models (LLMs) for autoregressive time series forecasting. By utilizing the generative capabilities of LLMs, the model can autoregressively predict future time steps based on past observations, capturing complex temporal patterns and dependencies. The Moment \cite{goswami2024moment} framework aims to provide a unified architecture for a wide range of time series tasks, including forecasting, classification, and anomaly detection. TIME-MOE \cite{shi2024time} introduces a scalable, expert-driven approach for time series forecasting using a mixture of experts (MoE) model. A unified approach, TIMER-XL \cite{liu2024timerxl}, is introduced, which enhances traditional transformer architectures by incorporating mechanisms designed to efficiently manage long-context information. This makes it particularly effective for forecasting tasks that require capturing long-term temporal dependencies. By integrating advanced attention mechanisms and optimized temporal representations, TimeMachine \cite{ahamed2024timemachine} greatly enhances the accuracy and reliability of long-term forecasts. A model-agnostic framework \cite{li2024functional} is introduced for multivariate time series forecasting that is independent of any specific model architecture. This framework focuses on learning and capturing the functional relationships between multiple time-dependent variables, allowing it to adapt to different forecasting tasks without being tied to any single underlying model.

\section{Methodology}
In this section, we have provided a comprehensive discussion of the key problem statement associated with time series forecasting and outlined the overall approach to addressing these challenges using state-of-the-art (SOTA) large-scale time-series models (LSTSMs) and small-scale time series transformers. Furthermore, we have presented a detailed empirical analysis comparing LSTSMs and small-scale transformers on a set of benchmark datasets, highlighting their differences in terms of performance, interpretability, and suitability for the time-series forecasting task.

\subsection{Problem Statement}
For a time series containing \( C \) variates, given historical samples \( X_{1:L} \) with a look-back window \( L \), each \( X_t \) at time step \( t \) is a vector of dimension \( C \). The time series forecasting task aims to predict the corresponding sequence \( X_{L+1:L+\tau} \) over the future \( \tau \) time steps. This task encourages a longer output horizon \( \tau \) to meet the demands of real-world scenarios.

\subsection{Proposed Workflow}
\begin{figure*}[hbt!]
\begin{center}
\includegraphics[scale=0.25]{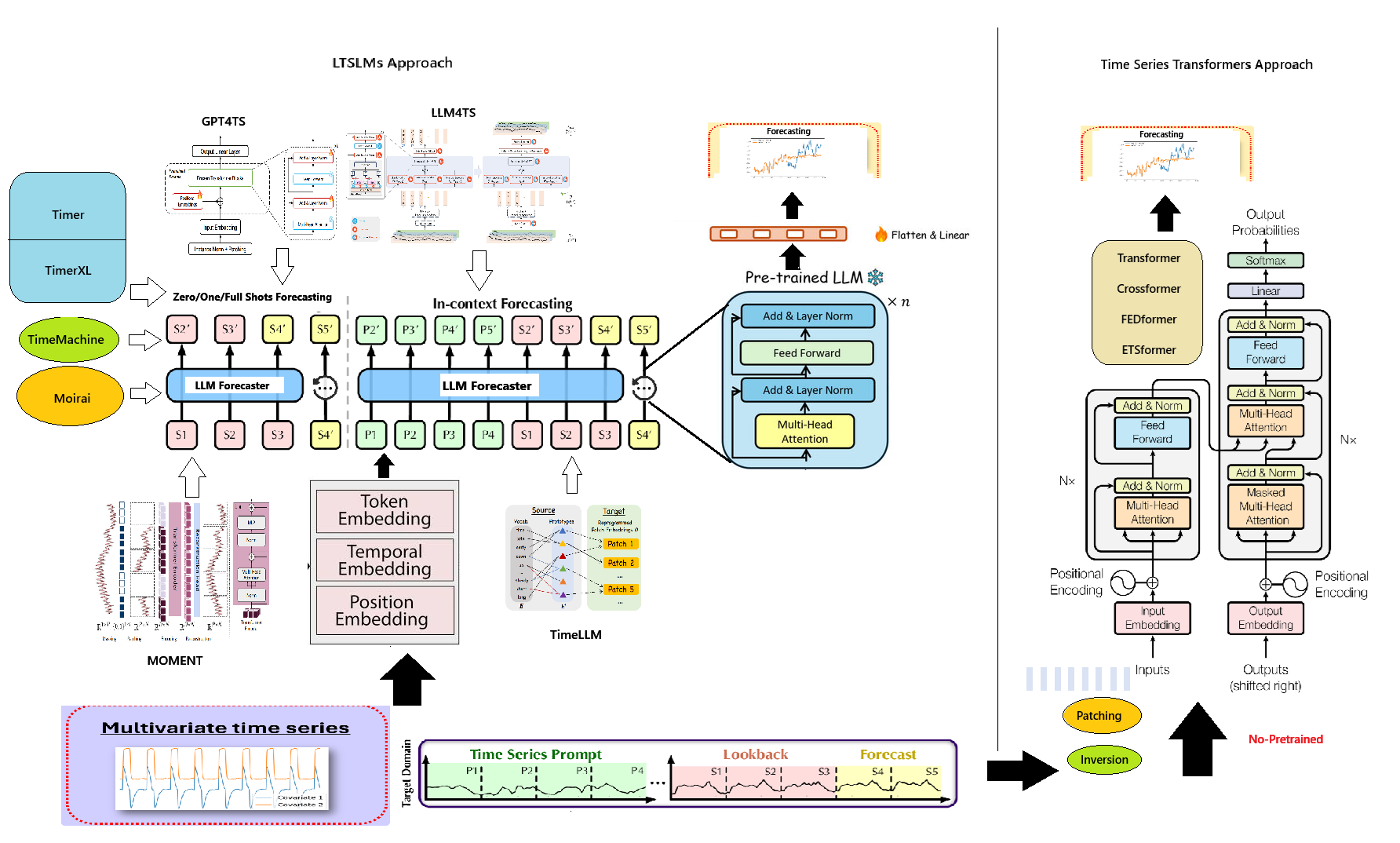}
\caption{Architectural Flow of the Proposed Study}
\label{Figarch}
\end{center}
\end{figure*}

The overall workflow of the proposed study is depicted in Figure \ref{Figarch}. Figure \ref{Figarch} distinguishes between the two approaches, LSTSMs and time series transformer models, while illustrating the general workflow for multivariate time series forecasting. In this study, we have presented a comprehensive comparison of two approaches for handling time series forecasting tasks. The first approach focuses on leveraging state-of-the-art (SOTA) pre-trained Large-scale Time Series Models (LSTSMs) specifically designed or adapted or fine-tuned for time series analysis tasks. In this paper, we have explored and analyzed a collection of LSTSMs, including Moment, Moirai, Timer, Timer\_XL, LLM4TS, GPT4TS, TimeMachine, and TimeLLM, focusing on their performance in time series forecasting tasks. We evaluate their performance under different fine-tuning strategies, including zero-shot learning (where the model is used as-is without additional training), one-shot learning (fine-tuned by a specific labeled data), few-shot learning (where a small amount of labeled data is provided for adaptation), and full-shot learning (where the model is fine-tuned extensively with task-specific data). This helps in assessing how well these pre-trained models generalize to time series forecasting. The second approach examines a different set of SOTA transformers that are not pre-trained on any specific time series corpus but instead are trained from scratch or fine-tuned on task-specific data. These models include Crossformer, ETSformer, iTransformer, PatchTST, and FEDformer. We investigate how these models learn temporal dependencies and patterns in time series data without leveraging prior knowledge from large-scale pretraining. By comparing the two approaches, we aim to analyze their respective strengths and weaknesses in terms of accuracy, computational efficiency, adaptability to new datasets, and explainability. This comparison provides insights into whether pre-trained LSTSMs offer a significant advantage over no-pretrained small-scale transformers or if training models from scratch remains a competitive and viable approach for time series forecasting.

For patch-based LSTSMs (such as Moment, Moirai etc,) patches are randomly masked in a uniform manner during pretraining by replacing their patch embeddings with a special mask embedding [MASK]. The objective of pretraining is to learn patch embeddings that enable the reconstruction of the input time series using a lightweight reconstruction head. The goal of the prediction head is to map the transformed patch embeddings to the desired output dimensions, effectively translating the learned representations into the final predicted values.

\section{Empirical Analysis}
\subsection{Pre-training and Execution Environment}
This paper utilizes datasets spanning long-term and short-term time series forecasting. All datasets used in this study are publicly available and are partitioned into training, validation, and test sets within the benchmark Time-Series Library (TSLib) \cite{sa_timeseries}. For pre-training LSTSMs, we have used a large dataset, such as the ERA5 family. We have curated a large-scale multivariate dataset, ERA5-Large, comprising 40 years of data from 4,920 stations, to develop a domain-specific model \cite{sa_timeseries}. The Moirai\_small model consists of 6 layers and 14 million parameters. We chose Moirai\_small for this experiment due to its simple architecture and relatively low resource demands, making it an efficient option for our analysis. The Moment\_small model consists of 12 layers and 125 million parameters. We selected the Moment\_small model with zero-shot fine-tuning for this experiment due to its simple architecture and lower resource requirements, which made it an efficient choice for the task. We have downloaded these datasets and necessary models from the 'Hugging Face' repository \cite{huggingface}. On the other hand, the GPT4TS, CALF-LLM and LLM4TS models leverage a pre-trained GPT-2 backbone for a wide range of tasks. The Time-LLM model uses Llama-7B \cite{touvron2023llama} as the default backbone in this experiment.
Every experiment is carried out on a single NVIDIA GeForce RTX 3090 GPU using PyTorch and CUDA version 12.2, and we have also used the Berzelius \cite{nsc_berzelius} high-performance computing (HPC) cluster platform for the large-scale models pre-training. Features 60 compute nodes, each equipped with 8 NVIDIA A100 GPUs (80GB), totaling 480 GPUs. The system is powered by AMD EPYC 7742 processors, providing 128 cores per node, and is interconnected via a 400 Gbps NVIDIA Mellanox HDR InfiniBand network for high-speed data transfer. Berzelius offers a total memory of 1.5 petabytes of NVMe storage and 94 TB of RAM. 

\subsection{Long-term Forecasting}
\begin{figure}[hbt!]
  \centering
  \begin{subfigure}{0.3\columnwidth}
    \includegraphics[width=\textwidth]{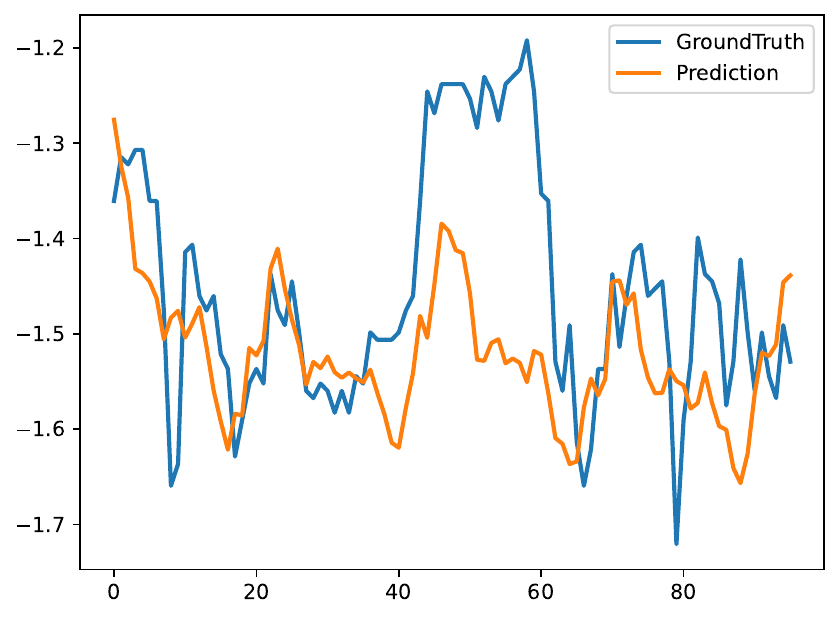}
    \caption{Timer} 
  \end{subfigure}
  \hfill 
  \begin{subfigure}{0.3\columnwidth}
    \includegraphics[width=\textwidth]{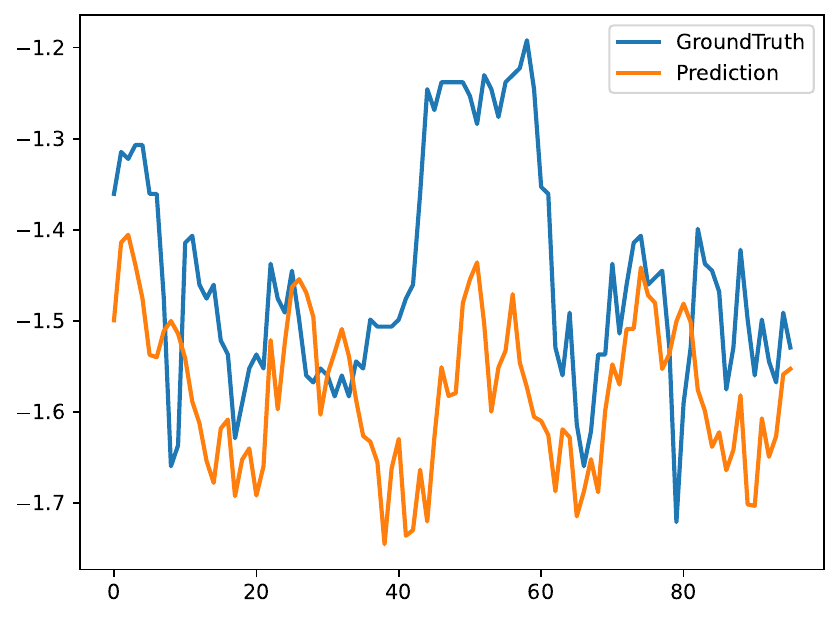}
    \caption{Moirai} 
  \end{subfigure}
  \hfill
  \begin{subfigure}{0.3\columnwidth}
    \includegraphics[width=\textwidth]{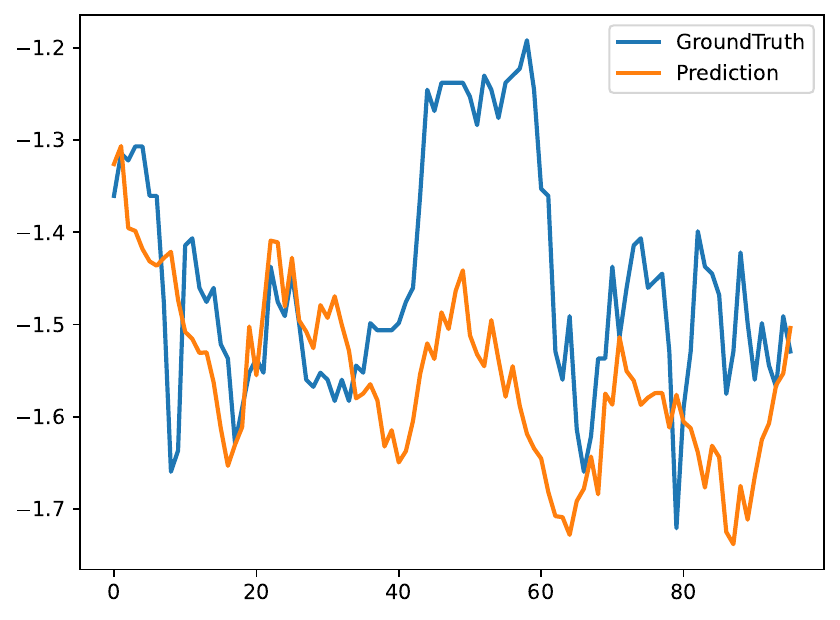}
    \caption{Moment} 
  \end{subfigure}
  
  \vspace{1em} 
  \begin{subfigure}{0.3\columnwidth}
    \includegraphics[width=\textwidth]{Moirai_ETTh1_96.pdf}
    \caption{Moirai} 
  \end{subfigure}
    \hfill 
  \begin{subfigure}{0.3\columnwidth}
    \includegraphics[width=\textwidth]{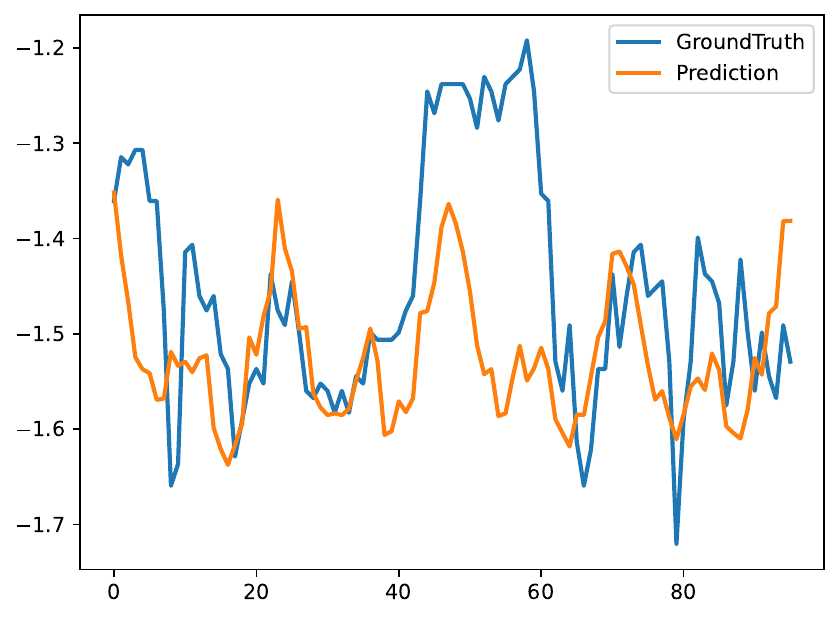}
    \caption{iTransformer} 
  \end{subfigure}
  \hfill
  \begin{subfigure}{0.3\columnwidth}
    \includegraphics[width=\textwidth]{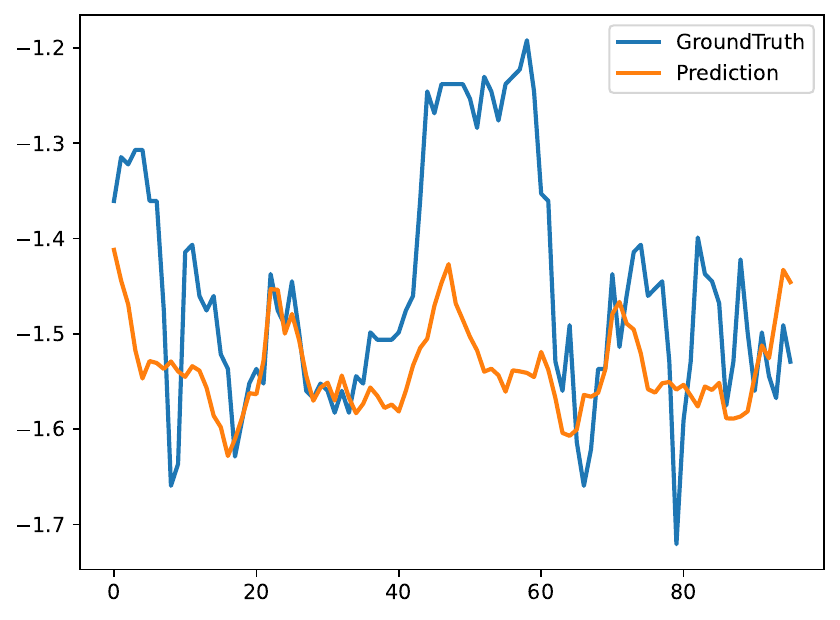}
    \caption{PatchTST} 
  \end{subfigure}

\vspace{0.5cm}

  \caption{Visualization of sample prediction (length: 96) results on the ETTh1 dataset for LSTSMs (training from scratch) and transformer models.}

  \vspace{0.5cm}
  \label{predtestetth1}
\end{figure}

In this section, we have done extensive experimental analysis and comparison in multivariate time series forecasting methods for both sets of LSTSMs and transformers. Table \ref{avglong1} presents a comparative analysis of state-of-the-art large-scale time series models (LSTSMs), including both models trained from scratch and non-pretrained transformers, across eight benchmark datasets. The performance is evaluated using average Mean Squared Error (MSE) and Mean Absolute Error (MAE) across four prediction lengths (96, 192, 336, 720) with a fixed context (lookback) length of 672. The Timer model achieves the lowest average MSE on multiple datasets, while Moment and GPT4TS show strong performance in terms of MAE.
The LLM-based models include variations like Timer, CALF, TimeLLM, TimeMachine, Moment, Moirai, LLM4TS, and GPT4TS, while the non-pretrained models include Crossformer, ETSFormer, iTransformer, PatchTST, and FEDformer. Table \ref{avglong2} summarizes the performance of state-of-the-art pretrained and fine-tuned large-scale time series models (LSTSMs), alongside non-pretrained transformers, on multivariate long-term forecasting across seven benchmark datasets. The evaluation is based on average MSE and MAE over various prediction lengths with a fixed lookback window (L = 672). Both full-shot (FS) and few-shot (FWS) training setups are considered. Among the models, LLM4TS\_FS achieves the best performance with the lowest MSE across most datasets and also yields the best MAE in several cases. GPT4TS and TimeMachine also show competitive results, particularly in Electricity and ETTh2 datasets, demonstrating the effectiveness of pretrained models in data-efficient forecasting scenarios. Comparing the two tables, the pretrained and fine-tuned model LLM4TS\_FS consistently outperforms others, achieving the best average MSE across four out of seven datasets (ETTh1, ETTh2, ETTm1, ETTm2) and the lowest average MAE in the same datasets, making it the overall top performer. While models like GPT4TS and Moment show competitive results in specific datasets, LLM4TS\_FS demonstrates superior generalization and accuracy, especially in long-term multivariate forecasting tasks.
Figure \ref{predtestetth1} illustrates the variation between model predictions and ground truth for selected LSTSMs and representative transformers in the case of long-term analysis of the ETTh1 dataset. Additional figures are provided in the supplementary file.

\begin{table*}[hbt!]
\scriptsize
\tiny
\caption{An overview of the benchmark datasets' outcomes for SOTA large-scale time series models (LSTSMs) (training from scratch) and SOTA no-pretrained transformers. Measure of average error coefficients on multivariate long-term forecasting results with different prediction lengths (96, 192, 336, 720) and context length C = 672. The lookback length is set as L = 672. The red color denotes the best average MSE, and the blue color denotes the best average MAE values.}
\begin{center}
\begin{tabular}{|c|c|cccccccc|}
\hline
\textbf{Model}                 & \textbf{Metric} & \textbf{ETTh1}               & \textbf{ETTh2}               & \textbf{ETTm1}               & \textbf{ETTm2}               & \textbf{Weather}             & \textbf{Electricity}         & \textbf{Traffic}             & \textbf{Exchange}            \\ \hline
                               & MSE             & {\color[HTML]{FE0000} 0.393} & 0.379                        & 0.345                        & {\color[HTML]{FE0000} 0.264} & 0.232                        & 0.183                        & 0.458                        & 0.454                        \\
\multirow{-2}{*}{Timer}        & MAE             & {\color[HTML]{3531FF} 0.418} & 0.417                        & 0.377                        & {\color[HTML]{3531FF} 0.321} & {\color[HTML]{3531FF} 0.268} & 0.276                        & 0.319                        & 0.453                        \\ \hline
                               & MSE             & 0.410                        & 0.373                        & 0.372                        & 0.275                        & 0.256                        & 0.188                        & 0.470                        & 0.494                        \\
\multirow{-2}{*}{Timer\_XL}    & MAE             & 0.433                        & 0.411                        & 0.392                        & 0.327                        & 0.291                        & 0.278                        & 0.344                        & 0.479                        \\ \hline
                               & MSE             & 0.415                        & 0.382                        & 0.348                        & {\color[HTML]{FE0000} 0.264} & {\color[HTML]{FE0000} 0.230} & {\color[HTML]{FE0000} 0.166} & 0.435                        & 0.450                        \\
\multirow{-2}{*}{Moment}       & MAE             & 0.439                        & 0.413                        & 0.385                        & 0.324                        & {\color[HTML]{6434FC} 0.268} & {\color[HTML]{343434} 0.266} & 0.303                        & 0.445                        \\ \hline
                               & MSE             & 0.500                        & 0.403                        & 0.370                        & 0.294                        & 0.248                        & 0.176                        & 0.448                        & 0.614                        \\
\multirow{-2}{*}{Moirai}       & MAE             & 0.488                        & 0.438                        & 0.400                        & 0.348                        & 0.287                        & 0.269                        & 0.310                        & 0.510                        \\ \hline
                               & MSE             & 0.428                        & {\color[HTML]{FE0000} 0.355} & 0.352                        & 0.267                        & 0.237                        & {\color[HTML]{FE0000} 0.166} & {\color[HTML]{FE0000} 0.414} & {\color[HTML]{343434} 0.424} \\
\multirow{-2}{*}{GPT4TS}       & MAE             & 0.426                        & {\color[HTML]{3531FF} 0.395} & 0.383                        & 0.326                        & 0.271                        & {\color[HTML]{3531FF} 0.263} & {\color[HTML]{3531FF} 0.295} & {\color[HTML]{343434} 0.426} \\ \hline
                               & MSE             & 0.557                        & 2.768                        & 0.591                        & 1.296                        & 0.265                        & 0.278                        & 0.563                        & 0.755                        \\
\multirow{-2}{*}{Crossformer}  & MAE             & 0.537                        & 1.324                        & 0.567                        & 0.719                        & 0.327                        & 0.340                        & 0.304                        & 0.645                        \\ \hline
                               & MSE             & 0.610                        & 0.441                        & {\color[HTML]{FE0000} 0.304} & 0.292                        & 0.263                        & 0.207                        & 0.620                        & {\color[HTML]{343434} 0.410} \\
\multirow{-2}{*}{ETSFormer}    & MAE             & 0.582                        & 0.455                        & {\color[HTML]{3531FF} 0.359} & 0.342                        & 0.319                        & 0.323                        & 0.395                        & 0.427                        \\ \hline
                               & MSE             & 0.457                        & 0.393                        & 0.406                        & 0.289                        & 0.252                        & 0.181                        & 0.444                        & {\color[HTML]{FE0000} 0.377} \\
\multirow{-2}{*}{iTransformer} & MAE             & 0.448                        & 0.411                        & 0.411                        & 0.332                        & 0.281                        & 0.270                        & 0.301                        & {\color[HTML]{3531FF} 0.418} \\ \hline
                               & MSE             & 0.457                        & 0.393                        & 0.365                        & 0.292                        & 0.257                        & 0.212                        & 0.532                        & 0.393                        \\
\multirow{-2}{*}{PatchTST}     & MAE             & 0.453                        & 0.415                        & 0.391                        & 0.334                        & 0.279                        & 0.309                        & 0.342                        & 0.418                        \\ \hline
                               & MSE             & 0.439                        & 0.442                        & 0.449                        & 0.307                        & 0.312                        & 0.295                        & 0.615                        & 0.520                        \\
\multirow{-2}{*}{FEDformer}    & MAE             & 0.458                        & 0.454                        & 0.457                        & 0.351                        & 0.364                        & 0.385                        & 0.383                        & 0.502                        \\ \hline
\end{tabular}
\label{avglong1}
\end{center}
\end{table*}

\begin{table*}[hbt!]
\scriptsize
\tiny
\caption{An overview of the benchmark datasets' outcomes for SOTA large-scale time series models (LSTSMs) (pre-trained and fine-tuning) and SOTA no-pretrained transformers. Measure of average error coefficients on multivariate long-term forecasting results with different prediction lengths (96, 192, 336, 720) and context length C = 672. The lookback length is set as L = 672. The red colour denotes the best average MSE, and the blue colour denotes the best average MAE values. FWS denotes a few-shot (on 50\% data), and FS denotes a full-shot.}
\begin{center}
\begin{tabular}{|c|c|ccccccc|}
\hline
\textbf{Model}                    & \textbf{Metric} & \textbf{ETTh1}               & \textbf{ETTh2}               & \textbf{ETTm1}               & \textbf{ETTm2}               & \textbf{Weather}             & \textbf{Electricity}         & \textbf{Exchange}            \\ \hline
                                  & MSE             & 0.510                        & 0.367                        & 0.400                        & 0.285                        & 0.243                        & 0.205                        & 0.381                        \\
\multirow{-2}{*}{Timer\_FS}       & MAE             & 0.508                        & 0.405                        & 0.412                        & 0.334                        & 0.279                        & 0.310                        & 0.419                        \\ \hline
                                  & MSE             & 0.712                        & 0.387                        & 0.506                        & 0.278                        & 0.252                        & 0.250                        & {\color[HTML]{FE0000} 0.314} \\
\multirow{-2}{*}{Timer\_FWS}      & MAE             & 0.585                        & 0.419                        & 0.464                        & 0.328                        & 0.286                        & 0.353                        & {\color[HTML]{3531FF} 0.384} \\ \hline
                                  & MSE             & 0.516                        & 0.374                        & 0.396                        & 0.282                        & 0.237                        & 0.186                        & 0.465                        \\
\multirow{-2}{*}{Timer\_XL\_FS}   & MAE             & 0.509                        & 0.417                        & 0.409                        & 0.334                        & 0.275                        & 0.288                        & 0.456                        \\ \hline
                                  & MSE             & 0.749                        & 0.481                        & 0.696                        & 0.276                        & 0.238                        & 0.190                        & 0.684                        \\
\multirow{-2}{*}{Timer\_XL\_FWS}  & MAE             & 0.617                        & 0.495                        & 0.543                        & 0.345                        & 0.282                        & 0.294                        & 0.542                        \\ \hline
                                  & MSE             & 0.460                        & 0.389                        & 0.410                        & 0.296                        & 0.274                        & 0.223                        & 0.562                        \\
\multirow{-2}{*}{TimeLLM\_FS}     & MAE             & 0.449                        & 0.408                        & 0.409                        & 0.340                        & 0.290                        & 0.309                        & 0.529                        \\ \hline
                                  & MSE             & 0.417                        & 0.344                        & 0.375                        & 0.268                        & 0.243                        & 0.170                        & 0.442                        \\
\multirow{-2}{*}{TimeMachine\_FS} & MAE             & 0.419                        & {\color[HTML]{3531FF} 0.383} & 0.392                        & 0.318                        & 0.273                        & {\color[HTML]{3531FF} 0.263} & 0.440                        \\ \hline
                                  & MSE             & 0.427                        & 0.354                        & 0.351                        & 0.266                        & 0.236                        & {\color[HTML]{FE0000} 0.166} & 0.424                        \\
\multirow{-2}{*}{GPT4TS\_FS}      & MAE             & 0.426                        & 0.394                        & 0.383                        & 0.326                        & 0.270                        & {\color[HTML]{3531FF} 0.263} & 0.436                        \\ \hline
                                  & MSE             & 0.518                        & 0.369                        & 0.421                        & 0.262                        & {\color[HTML]{FE0000} 0.232} & 0.174                        & 0.377                        \\
\multirow{-2}{*}{GPT4TS\_FWS}     & MAE             & 0.498                        & 0.412                        & 0.417                        & 0.320                        & {\color[HTML]{3531FF} 0.269} & 0.273                        & 0.446                        \\ \hline
                                  & MSE             & {\color[HTML]{FE0000} 0.404} & {\color[HTML]{FE0000} 0.333} & {\color[HTML]{FE0000} 0.343} & {\color[HTML]{FE0000} 0.251} & 0.233                        & 0.169                        & 0.415                        \\
\multirow{-2}{*}{LLM4TS\_FS}      & MAE             & {\color[HTML]{3531FF} 0.418} & {\color[HTML]{3531FF} 0.383} & {\color[HTML]{3531FF} 0.378} & {\color[HTML]{3531FF} 0.313} & 0.270                        & 0.265                        & 0.424                        \\ \hline
\end{tabular}
\label{avglong2}
\end{center}
\end{table*}

\subsection{Short-term Forecasting}
\begin{figure}[hbt!]
  \centering
  \begin{subfigure}{0.3\columnwidth}
    \includegraphics[width=\textwidth]{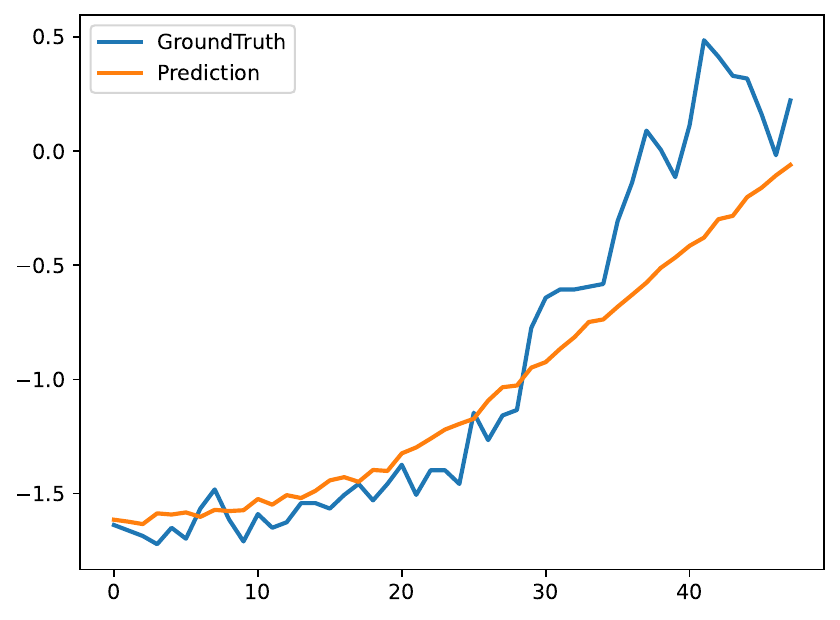}
    \caption{Moment} 
  \end{subfigure}
  \hfill 
  \begin{subfigure}{0.3\columnwidth}
    \includegraphics[width=\textwidth]{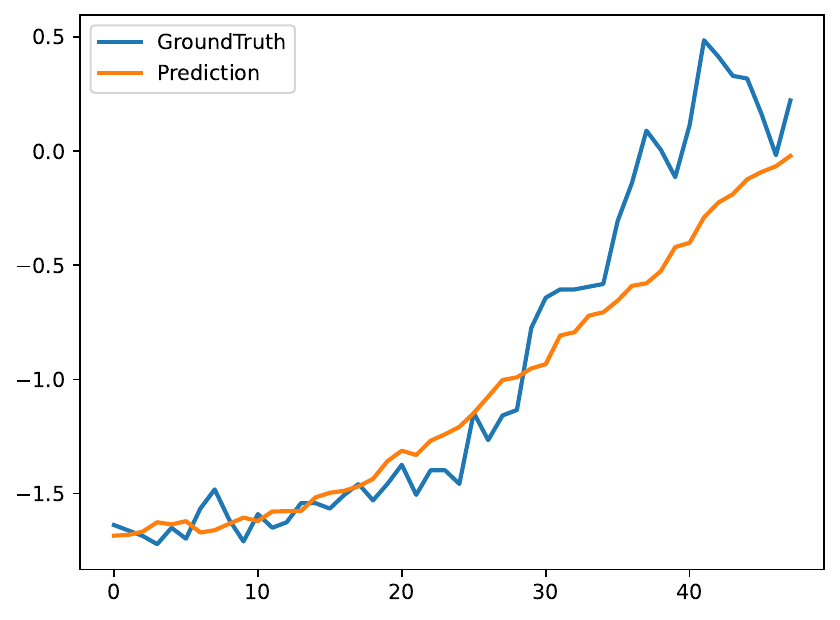}
    \caption{Moirai} 
  \end{subfigure}
  \hfill
  \begin{subfigure}{0.3\columnwidth}
    \includegraphics[width=\textwidth]{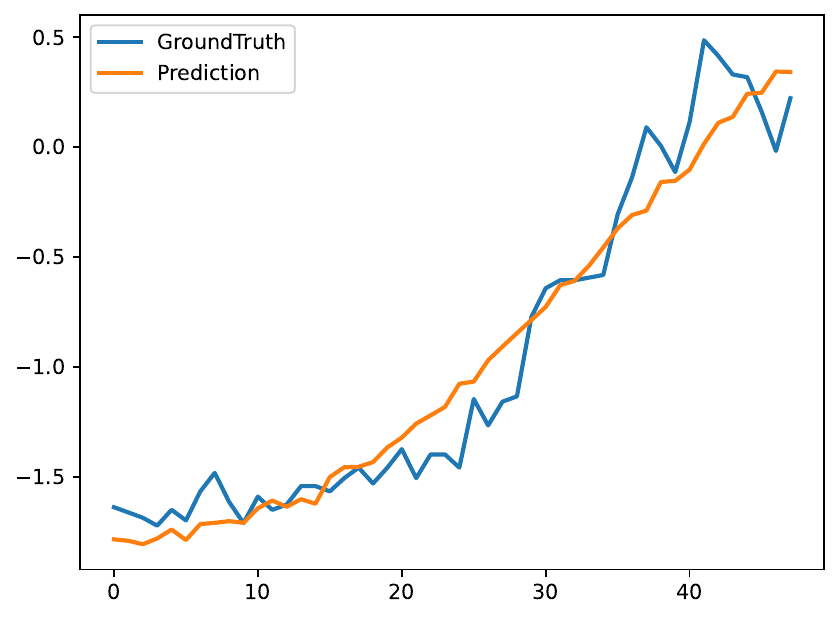}
    \caption{Timer} 
  \end{subfigure}
  
  \vspace{1em} 
  \begin{subfigure}{0.3\columnwidth}
    \includegraphics[width=\textwidth]{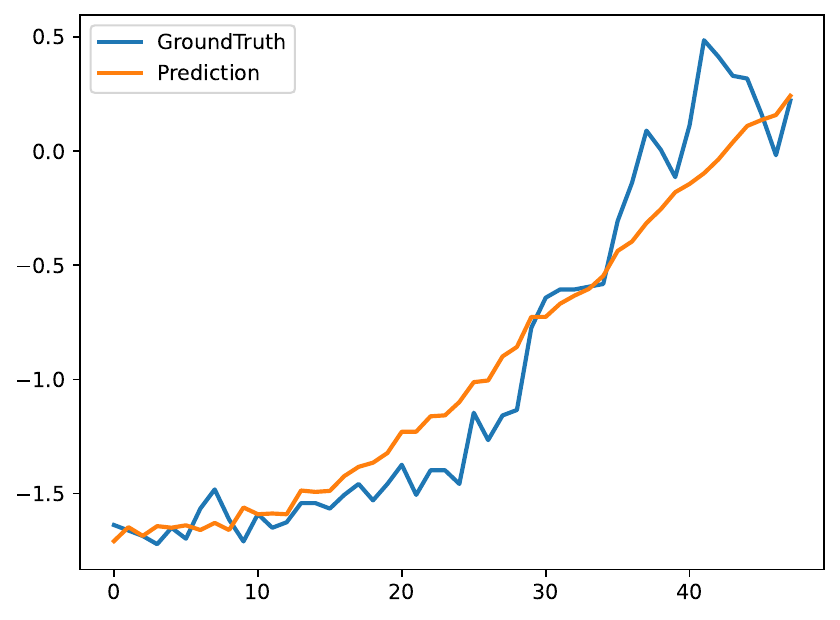}
    \caption{Timer\_XL} 
  \end{subfigure}
    \hfill 
  \begin{subfigure}{0.3\columnwidth}
    \includegraphics[width=\textwidth]{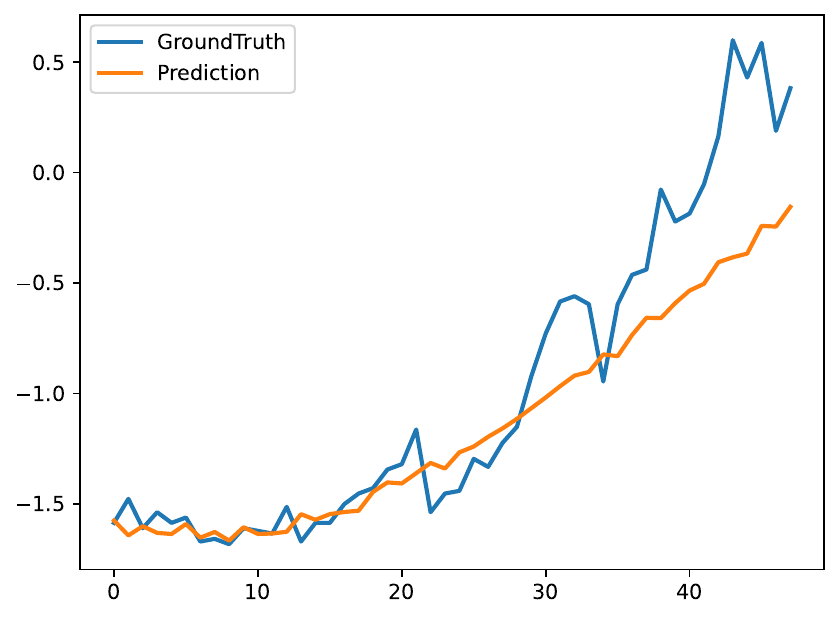}
    \caption{iTransformer} 
  \end{subfigure}
  \hfill
  \begin{subfigure}{0.3\columnwidth}
    \includegraphics[width=\textwidth]{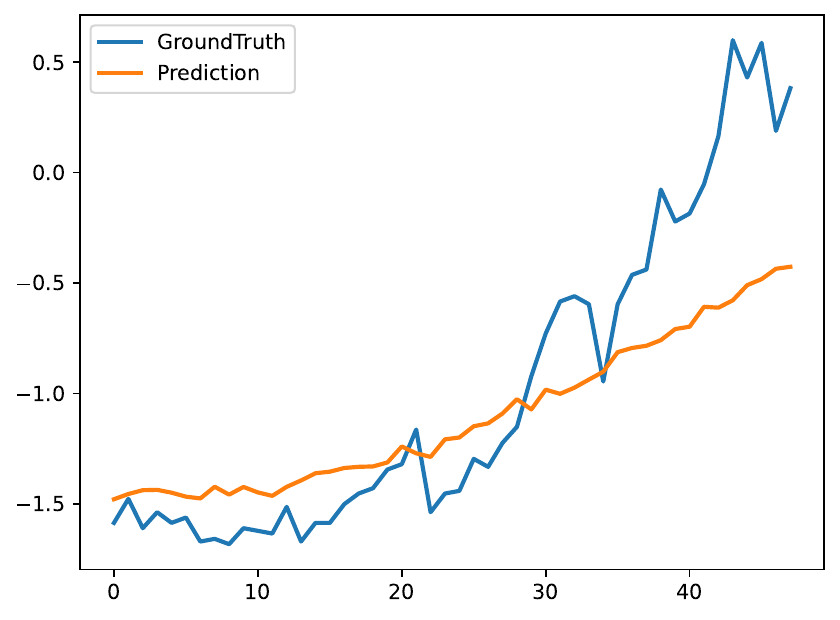}
    \caption{PatchTST} 
  \end{subfigure}
  
   \vspace{0.5cm} 
  \caption{Visualization of sample prediction results of LSTSMs and transformers on PEMS04 dataset for prediction length 48.}
  \label{fig-PEMS04-compare}
   \vspace{0.5cm} 
\end{figure}

For short-term forecasting analysis, we have used two benchmark datasets: the PEMS dataset and the M4 dataset. Table \ref{tabPEMS} presents a comparative analysis of error coefficients (MSE and MAE) for a set of SOTA LSTSMs and small-scale transformers on multivariate short-term forecasting tasks across different prediction horizons (12, 24, and 48). The evaluation is conducted on three traffic datasets (PEMS03, PEMS04, and PEMS08) with a learning rate of 0.0001, five training epochs, and a fixed look-back window of 96. The red-highlighted values indicate the lowest MSE, while the blue-highlighted values represent the lowest MAE, signifying the best-performing models for each scenario. Results show that Timer\_XL consistently achieves optimal performance across multiple settings, followed closely by Crossformer and Moirai. Timer\_XL outperforms other models in multivariate short-term forecasting by effectively capturing temporal dependencies over long sequences. Its superior performance, as indicated by the lowest MSE and MAE values in several cases, stems from its enhanced attention mechanisms and efficient architectural design, which balance complexity and computational efficiency. Unlike other models, Timer\_XL leverages optimized feature extraction, making it more robust to varying time series patterns and improving generalization across datasets.  Figure \ref{fig-PEMS04-compare} illustrates the variation between model predictions and ground truth for selected LSTSMs and representative transformers in the case of short-term analysis of the PEMS04 dataset. It is important to note that the architecture of the underlying execution environment influences this comparison.

\begin{table*}[hbt!]
\scriptsize
\tiny
\caption{Comparison of error coefficients on multivariate short-term forecasting results on PEMS datasets with prediction horizons (12, 24, 48), learning\_rate 0.0001, and fixed look-back 96. The red and blue highlighted colour values represent the best MSE and MAE, respectively.}
\begin{center}
\begin{tabular}{|c|c|ll|ll|ll|ll|ll|ll|ll|}
\hline
\multicolumn{1}{|l|}{}                                     & Models   & \multicolumn{2}{c|}{Moment}                             & \multicolumn{2}{c|}{Moirai}                             & \multicolumn{2}{c|}{Timer}                              & \multicolumn{2}{c|}{Timer\_XL}                                                                        & \multicolumn{2}{c|}{Crossformer}                                                                      & \multicolumn{2}{c|}{PatchTST}                           & \multicolumn{2}{c|}{iTransformer}                       \\ \cline{2-16} 
\multicolumn{1}{|l|}{\multirow{-2}{*}{Prediction\_Length}} & Database & \multicolumn{1}{c}{MSE}    & \multicolumn{1}{c|}{MAE}   & \multicolumn{1}{c}{MSE}    & \multicolumn{1}{c|}{MAE}   & \multicolumn{1}{c|}{MSE}   & \multicolumn{1}{c|}{MAE}   & \multicolumn{1}{c|}{MSE}                          & \multicolumn{1}{c|}{MAE}                          & \multicolumn{1}{c}{MSE}                           & \multicolumn{1}{c|}{MAE}                          & \multicolumn{1}{c}{MSE}    & \multicolumn{1}{c|}{MAE}   & \multicolumn{1}{c}{MSE}    & \multicolumn{1}{c|}{MAE}   \\ \hline
                                                           & PEMS03   & \multicolumn{1}{l|}{0.094} & 0.208                      & \multicolumn{1}{l|}{0.078} & 0.185                      & \multicolumn{1}{l|}{0.091} & 0.201                      & \multicolumn{1}{l|}{0.074}                        & 0.182                                             & \multicolumn{1}{l|}{{\color[HTML]{FE0000} 0.072}} & {\color[HTML]{3531FF} 0.178}                      & \multicolumn{1}{l|}{0.099} & 0.214                      & \multicolumn{1}{l|}{0.079} & 0.188                      \\ \cline{2-16} 
                                                           & PEMS04   & \multicolumn{1}{l|}{0.113} & 0.232                      & \multicolumn{1}{l|}{0.102} & 0.212                      & \multicolumn{1}{l|}{0.112} & 0.223                      & \multicolumn{1}{l|}{{\color[HTML]{FE0000} 0.091}} & {\color[HTML]{3531FF} 0.198}                      & \multicolumn{1}{l|}{0.092}                        & 0.199                                             & \multicolumn{1}{l|}{0.115} & 0.229                      & \multicolumn{1}{l|}{0.095} & 0.203                      \\ \cline{2-16} 
\multirow{-3}{*}{12}                                       & PEMS08   & \multicolumn{1}{l|}{0.103} & 0.221                      & \multicolumn{1}{l|}{0.089} & 0.197                      & \multicolumn{1}{l|}{0.100} & 0.211                      & \multicolumn{1}{l|}{{\color[HTML]{FE0000} 0.082}} & {\color[HTML]{3531FF} 0.189}                      & \multicolumn{1}{l|}{0.167}                        & 0.231                                             & \multicolumn{1}{l|}{0.102} & 0.216                      & \multicolumn{1}{l|}{0.099} & 0.211                      \\ \hline
                                                           & PEMS03   & \multicolumn{1}{c|}{0.160} & \multicolumn{1}{c|}{0.274} & \multicolumn{1}{c|}{0.118} & \multicolumn{1}{c|}{0.231} & \multicolumn{1}{c|}{0.154} & \multicolumn{1}{c|}{0.262} & \multicolumn{1}{c|}{{\color[HTML]{FE0000} 0.111}} & \multicolumn{1}{c|}{{\color[HTML]{3531FF} 0.225}} & \multicolumn{1}{c|}{0.153}                        & \multicolumn{1}{c|}{0.266}                        & \multicolumn{1}{c|}{0.162} & \multicolumn{1}{c|}{0.281} & \multicolumn{1}{c|}{0.120} & \multicolumn{1}{c|}{0.235} \\ \cline{3-16} 
                                                           & PEMS04   & \multicolumn{1}{c|}{0.188} & \multicolumn{1}{c|}{0.300} & \multicolumn{1}{c|}{0.148} & \multicolumn{1}{c|}{0.258} & \multicolumn{1}{c|}{0.188} & \multicolumn{1}{c|}{0.292} & \multicolumn{1}{c|}{{\color[HTML]{FE0000} 0.137}} & \multicolumn{1}{c|}{{\color[HTML]{3531FF} 0.248}} & \multicolumn{1}{c|}{{\color[HTML]{FE0000} 0.137}} & \multicolumn{1}{c|}{0.255}                        & \multicolumn{1}{c|}{0.221} & \multicolumn{1}{c|}{0.316} & \multicolumn{1}{c|}{0.145} & \multicolumn{1}{c|}{0.255} \\ \cline{3-16} 
\multirow{-3}{*}{24}                                       & PEMS08   & \multicolumn{1}{c|}{0.169} & \multicolumn{1}{c|}{0.284} & \multicolumn{1}{c|}{0.136} & \multicolumn{1}{c|}{0.249} & \multicolumn{1}{c|}{0.164} & \multicolumn{1}{c|}{0.273} & \multicolumn{1}{c|}{{\color[HTML]{FE0000} 0.125}} & \multicolumn{1}{c|}{{\color[HTML]{343434} 0.237}} & \multicolumn{1}{c|}{0.169}                        & \multicolumn{1}{c|}{{\color[HTML]{3531FF} 0.216}} & \multicolumn{1}{c|}{0.184} & \multicolumn{1}{c|}{0.297} & \multicolumn{1}{c|}{0.137} & \multicolumn{1}{c|}{0.244} \\ \hline
                                                           & PEMS03   & \multicolumn{1}{l|}{0.299} & 0.376                      & \multicolumn{1}{l|}{0.208} & 0.315                      & \multicolumn{1}{l|}{0.297} & 0.373                      & \multicolumn{1}{l|}{{\color[HTML]{FE0000} 0.200}} & {\color[HTML]{3531FF} 0.308}                      & \multicolumn{1}{l|}{0.287}                        & 0.393                                             & \multicolumn{1}{l|}{0.301} & 0.379                      & \multicolumn{1}{l|}{0.210} & 0.320                      \\ \cline{2-16} 
                                                           & PEMS04   & \multicolumn{1}{l|}{0.369} & 0.429                      & \multicolumn{1}{l|}{0.258} & 0.350                      & \multicolumn{1}{l|}{0.374} & 0.425                      & \multicolumn{1}{l|}{{\color[HTML]{FE0000} 0.241}} & {\color[HTML]{3531FF} 0.337}                      & \multicolumn{1}{l|}{0.378}                        & 0.465                                             & \multicolumn{1}{l|}{0.382} & 0.449                      & \multicolumn{1}{l|}{0.255} & 0.349                      \\ \cline{2-16} 
\multirow{-3}{*}{48}                                       & PEMS08   & \multicolumn{1}{l|}{0.335} & 0.410                      & \multicolumn{1}{l|}{0.250} & 0.349                      & \multicolumn{1}{l|}{0.336} & 0.404                      & \multicolumn{1}{l|}{{\color[HTML]{FE0000} 0.235}} & {\color[HTML]{3531FF} 0.337}                      & \multicolumn{1}{l|}{0.310}                        & 0.401                                             & \multicolumn{1}{l|}{0.358} & 0.419                      & \multicolumn{1}{l|}{0.259} & 0.345                      \\ \hline
\end{tabular}
\end{center}
\label{tabPEMS}
\end{table*}

\subsection{Quality Comparison}
Table \ref{qualitycomp} presents a comparative analysis of state-of-the-art (SOTA) large-scale time-series models (LSTSMs) and small-scale transformers for time series forecasting. The comparison highlights key aspects such as model architecture, size, pre-training scale, tokenization strategy, and typical applications. LSTSMs, including Timer, GPT4TS, LLM4TS, Timer\_xl, and Moment, generally have larger model sizes and extensive pre-training on billions of tokens, enabling them to handle multiple tasks like forecasting, classification, anomaly detection, and imputation. In contrast, small-scale transformers, such as Crossformer, PatchTST, and ETSformer, rely on encoder-based architectures with significantly smaller model sizes and no pre-training, focusing primarily on forecasting tasks. The choice between these models depends on factors like task complexity, computational efficiency, and the need for adaptability across different time series datasets.

\subsection{Analysis of Alpha Values for Pretraining and Finetuning}
The $\alpha$ metric, computed using the \textit{WeightWatcher} library, provides insight into the quality of learned representations in deep neural networks \citep{martin2021implicit}. It is derived from the spectral analysis of weight matrices and serves as a measure of how well a layer is trained. Specifically, an optimal $\alpha$ value typically falls within the range of \textit{2} to \textit{6}, indicating a well-trained and generalizable layer \citep{martin2021predicting}. Lower values may suggest underfitting, while excessively high values may indicate overfitting.

\begin{table}[hbt!]
\scriptsize
\tiny
\caption{Quality comparison of SOTA LSTSMs and small-scale Transformers for Time Series Analysis. M denotes million, T denotes trillion, and B denotes billion.}
\begin{center}
\begin{tabular}{|l|c|p{1cm}|p{1cm}|c|p{1.2cm}|}
\hline
\textbf{Model} & \textbf{Architecture} & \textbf{Size} & \textbf{Pre-training Scale} & \textbf{Token Type} & \textbf{Tasks} \\ \hline
Timer & Decoder & 29M, 50M, 67M & 28B & Patch & Forecasting, Imputation, Anomaly Detection \\ \hline
Timer\_XL & Decoder & 14M, 91M, 311M & 32B & Patch & Forecasting, Imputation, Anomaly Detection \\ \hline
LLama & Decoder & 200M & 0.36B & Time points & Forecasting \\ \hline
Chronos & Encoder+Decoder & 20M, 46M, 200M, 710M & 84B & Time points & Forecasting \\ \hline
TimeMachine & Encoder & 10M-100M & - & Time points & Forecasting \\ \hline
Moment & T5 Encoder & 40M, 125M, 385M & - & Patch & Forecasting, Classification, Anomaly Detection \\ \hline
Moirai & Encoder & 14M, 91M, 311M & - & Patch & Forecasting \\ \hline
LLM4TS & Decoder & 117M-345M & 1.5B & Patch & Forecasting \\ \hline
GPT4TS & Decoder & 1.7T & 175B & Patch & Forecasting, Classification, Anomaly Detection \\ \hline
Crossformer & Encoder+Decoder & 20M & - & Time points & Forecasting \\ \hline
ETSFormer & Encoder+Decoder & 10M & - & Time points & Forecasting \\ \hline
iTransformer & Encoder & 10M & - & Features & Forecasting \\ \hline
PatchTST & Encoder & 20M & - & Patch & Forecasting \\ \hline
FEDformer & Encoder+Decoder & 33M & - & Time points & Forecasting \\ \hline
\end{tabular}
\label{qualitycomp}
\end{center}
\end{table}

\begin{figure}[hbt!]
\begin{center}
\includegraphics[scale=0.27]{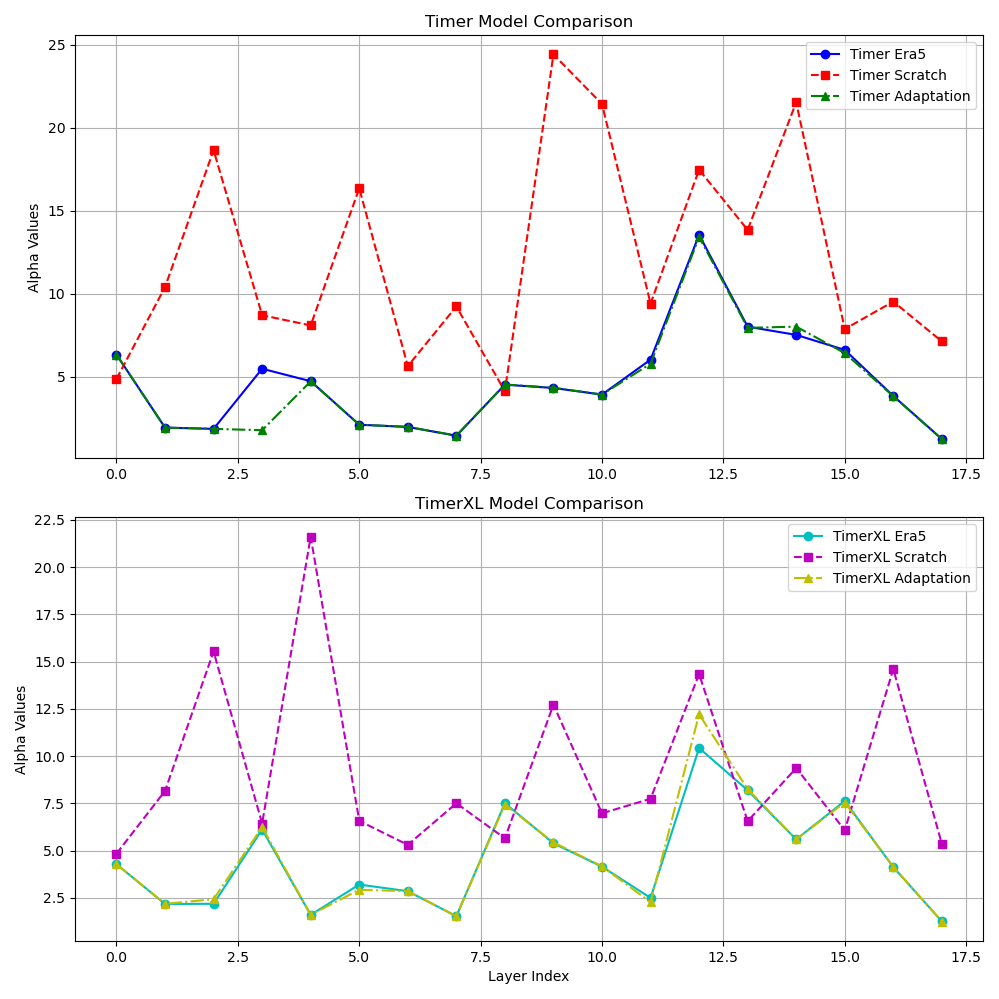}
\caption{Comparison of Alpha Values Across Different Training Strategies for Timer and Timer\_XL Models}
\vspace{5pt}
\label{fig-alpha}
\end{center}
\end{figure}

In this study, we compare the $\alpha$ values across different training approaches for the Timer and Timer\_XL models. These approaches include:
\begin{itemize}
\item \textbf{Pretraining on Era5:} The model is trained on a large-scale dataset (Era5) to establish strong initial representations.
\item \textbf{Training from scratch on ETTh1:} The model is trained directly on the ETTh1 dataset without pretraining.
\item\textbf{Adaptation from Era5 to ETTh1:} The model is first pretrained on Era5 and then fine-tuned on ETTh1 to leverage prior knowledge.
\end{itemize}
The primary objective of this analysis is to compare the learnable performance of different training strategies by evaluating their corresponding $\alpha$ values. Figure \ref{fig-alpha} demonstrates that models benefiting from pretraining on Era5 generally exhibit more stable $\alpha$ values within the optimal range, confirming the advantage of transfer learning in improving layer quality and generalization.

\section{Discussion}
\textbf{Theorem 1:} In multivariate long-term time series forecasting, pretrained and fine-tuned models minimize the expected loss \( \mathbb{E}[ L(f(X), Y)] \) more effectively than models trained from scratch, where \( L \) represents MSE (Mean Squared Error). Empirical evidence shows that \texttt{LLM4TS\_FS} achieves the lowest average MSE and MAE across multiple datasets, demonstrating superior generalization. Its performance highlights the effectiveness of leveraging pretraining and transfer learning to enhance forecasting accuracy and capture complex temporal dependencies across diverse domains.

\textbf{Theorem 2:} Let \(\mathcal{M}_p\) and \(\mathcal{M}_t\) represent the sets of Large-scale Time-Series Models (LSTSMs) and small-scale transformers, respectively. The generalization performance of LSTSMs is superior to that of small-scale transformers, as 
$\mathbb{E}[\mathcal{L}(f_p(X))] \leq \mathbb{E}[\mathcal{L}(f_t(X))]$,
where \( f_p \in \mathcal{M}_p \) and \( f_t \in \mathcal{M}_t \), assuming the benefit of large-scale pretraining for LSTSMs. Computationally, LSTSMs scale as \( \mathcal{O}(T \cdot d) + \mathcal{O}(H) \), where \( H \) is the pretraining memory overhead, while small-scale transformers exhibit quadratic complexity \( \mathcal{O}(T^2 \cdot d) \), making them computationally expensive for long time-series. Furthermore, LSTSMs optimize multiple objectives (forecasting, classification, and anomaly detection), making them more adaptable for multi-task scenarios, whereas small-scale transformers focus solely on forecasting, serving as lightweight alternatives.

\textbf{Theorem 3:} In multivariate short-term traffic forecasting, let the loss function \( L \) be defined as MSE (Mean Squared Error). Then, the optimal model \( f^* \) minimizes \( \mathbb{E}[L(f(X), Y)] \) across varying prediction horizons and datasets. Empirical evaluation across PEMS03, PEMS04, and PEMS08 datasets with horizons of 12, 24, and 48 reveals that \texttt{Timer\_XL} consistently achieves the lowest error values in both MSE and MAE. Its performance indicates strong generalization and precise short-term forecasting, attributable to its effective attention mechanism and temporal feature learning.

\textbf{Theorem 4:} Let $\mathcal{M}_{\text{small}}$ denote the class of small-scale, task-specific time series transformer models, and $\mathcal{M}_{\text{large}}$ denote the class of large-scale, pretrained transformer models (e.g., \texttt{LLM4TS\_FS}, \texttt{Timer\_XL}). Let $\mathcal{D} = \{(X_t, Y_t)\}_{t=1}^T$ represent a multivariate time series dataset. Suppose both $\mathcal{M}_{\text{small}}$ and $\mathcal{M}_{\text{large}}$ are trained on $\mathcal{D}_{\text{train}} \subset \mathcal{D}$ and evaluated on $\mathcal{D}_{\text{test}}$.
Assume the expected loss over a prediction horizon $H$ is given by:
$\mathbb{E}_{(X_t, Y_t) \sim \mathcal{D}_{\text{test}}} \left[ \mathcal{L}(f(X_t; \theta), Y_{t+1:t+H}) \right]$
where $\mathcal{L} \in \{\text{MSE}, \text{MAE}\}$ and $f(\cdot; \theta)$ is a forecasting model parameterized by $\theta$. Then, under sufficient training data and model capacity, the following inequality holds:
$\min_{\theta \in \mathcal{M}_{\text{large}}} \mathbb{E}[\mathcal{L}] < \min_{\theta \in \mathcal{M}_{\text{small}}} \mathbb{E}[\mathcal{L}]$.

\textbf{Theorem 5:} Even without pretraining, large-scale transformer architectures $\mathcal{M}_{\text{scratch}} \subset \mathcal{M}_{\text{large}}$ (e.g., \texttt{Moment}, \texttt{Timer}) achieve competitive performance:
$\min_{\theta \in \mathcal{M}_{\text{scratch}}} \mathbb{E}[\mathcal{L}] \leq \min_{\theta \in \mathcal{M}_{\text{small}}} \mathbb{E}[\mathcal{L}]$
provided sufficient training data and optimization resources.

\section{Conclusions and Future Work}
This study provides a distinctive discussion on the empirical and theoretical insights into the competitiveness of state-of-the-art (SOTA) large-scale and small-scale time series transformer models for forecasting tasks. It encompasses both long-term and short-term forecasting to highlight the performance diversity of these models. The empirical and theoretical analyses across short-term, long-term, and multivariate time series forecasting tasks consistently demonstrate that pretrained large-scale transformer models, such as \texttt{LLM4TS\_FS} and \texttt{Timer\_XL}, significantly outperform smaller, task-specific models by minimizing expected loss functions—particularly MSE and MAE—across diverse datasets and prediction horizons. Their superior capacity for temporal pattern recognition, effective feature extraction via attention mechanisms, and enhanced generalization through transfer learning contribute to their state-of-the-art forecasting performance. Notably, large-scale models trained from scratch, such as \texttt{Moment} and \texttt{Timer}, also exhibit competitive results, highlighting the strength of scalable architectures even without pretraining, especially when equipped with advanced attention mechanisms and sufficient training data. This study aids future researchers in making informed decisions regarding model adaptation for various time series tasks. A limitation of this study is the lack of analysis on model interpretability and computational efficiency during inference for real-time deployment scenarios.
We have highlighted several promising directions for future work, including applying LSTSMs to real-world time series forecasting challenges, exploring multi-modal time series and text foundation models, and improving forecasting performance through strong pretraining with causal attention and forecasting objectives. More efficient tuning techniques, such as µP \cite{yang2022tensor}, can be leveraged. Future research will explore more flexible approaches for time series analysis, including classification, anomaly detection, and imputation, by incorporating advanced LSTSMs. 







\bibliography{mybibfile}

\begin{thebibliography}{48}
\providecommand{\natexlab}[1]{#1}
\providecommand{\url}[1]{\texttt{#1}}
\expandafter\ifx\csname urlstyle\endcsname\relax
  \providecommand{\doi}[1]{doi: #1}\else
  \providecommand{\doi}{doi: \begingroup \urlstyle{rm}\Url}\fi

\bibitem[Abdel-Sater and Ben~Hamza(2024)]{abdel2024federated}
R.~Abdel-Sater and A.~Ben~Hamza.
\newblock A federated large language model for long-term time series forecasting.
\newblock In \emph{ECAI 2024}, pages 2452--2459. IOS Press, 2024.

\bibitem[Ahamed and Cheng(2024)]{ahamed2024timemachine}
M.~A. Ahamed and Q.~Cheng.
\newblock Timemachine: A time series is worth 4 mambas for long-term forecasting.
\newblock In \emph{ECAI 2024}, volume 392, page 1688, 2024.

\bibitem[Ahmed et~al.(2023)Ahmed, Nielsen, Tripathi, Siddiqui, Ramachandran, and Rasool]{ahmed2023transformers}
S.~Ahmed, I.~E. Nielsen, A.~Tripathi, S.~Siddiqui, R.~P. Ramachandran, and G.~Rasool.
\newblock Transformers in time-series analysis: A tutorial.
\newblock \emph{Circuits, Systems, and Signal Processing}, 42\penalty0 (12):\penalty0 7433--7466, 2023.

\bibitem[Ansari et~al.(2024)Ansari, Stella, Turkmen, Zhang, Mercado, Shen, Shchur, Rangapuram, Arango, Kapoor, et~al.]{ansari2024chronos}
A.~F. Ansari, L.~Stella, C.~Turkmen, X.~Zhang, P.~Mercado, H.~Shen, O.~Shchur, S.~Rangapuram, S.~P. Arango, S.~Kapoor, et~al.
\newblock Chronos: Learning the language of time series.
\newblock 2024.

\bibitem[Campagner et~al.(2024)Campagner, Barandas, Folgado, Gamboa, and Cabitza]{campagner2024ensemble}
A.~Campagner, M.~Barandas, D.~Folgado, H.~Gamboa, and F.~Cabitza.
\newblock Ensemble predictors: Possibilistic combination of conformal predictors for multivariate time series classification.
\newblock \emph{IEEE Transactions on Pattern Analysis and Machine Intelligence}, 2024.

\bibitem[Chang et~al.(2023)Chang, Wang, Peng, and Chen]{chang2023llm4ts}
C.~Chang, W.-Y. Wang, W.-C. Peng, and T.-F. Chen.
\newblock Llm4ts: Aligning pre-trained llms as data-efficient time-series forecasters.
\newblock \emph{arXiv preprint arXiv:2308.08469}, 2023.

\bibitem[Das et~al.(2024)Das, Kong, Sen, and Zhou]{das2024decoder}
A.~Das, W.~Kong, R.~Sen, and Y.~Zhou.
\newblock A decoder-only foundation model for time-series forecasting.
\newblock In \emph{Forty-first International Conference on Machine Learning}, 2024.

\bibitem[Engel and Engel(2024)]{engel2024transformer}
E.~A. Engel and N.~E. Engel.
\newblock A transformer with a fuzzy attention mechanism for weather time series forecasting.
\newblock In \emph{International Conference on Neuroinformatics}, pages 418--425. Springer, 2024.

\bibitem[Face(2023)]{huggingface}
H.~Face.
\newblock Hugging face’s model hub, 2023.
\newblock URL \url{https://huggingface.co}.

\bibitem[Foumani et~al.(2024)Foumani, Tan, Webb, and Salehi]{foumani2024improving}
N.~M. Foumani, C.~W. Tan, G.~I. Webb, and M.~Salehi.
\newblock Improving position encoding of transformers for multivariate time series classification.
\newblock \emph{Data mining and knowledge discovery}, 38\penalty0 (1):\penalty0 22--48, 2024.

\bibitem[Goswami et~al.(2024)Goswami, Szafer, Choudhry, Cai, Li, and Dubrawski]{goswami2024moment}
M.~Goswami, K.~Szafer, A.~Choudhry, Y.~Cai, S.~Li, and A.~Dubrawski.
\newblock Moment: A family of open time-series foundation models.
\newblock \emph{arXiv preprint arXiv:2402.03885}, 2024.

\bibitem[Jin et~al.(2023)Jin, Wang, Ma, Chu, Zhang, Shi, Chen, Liang, Li, Pan, et~al.]{jin2023time}
M.~Jin, S.~Wang, L.~Ma, Z.~Chu, J.~Y. Zhang, X.~Shi, P.-Y. Chen, Y.~Liang, Y.-F. Li, S.~Pan, et~al.
\newblock Time-llm: Time series forecasting by reprogramming large language models.
\newblock \emph{arXiv preprint arXiv:2310.01728}, 2023.

\bibitem[Kitaev et~al.(2020)Kitaev, Kaiser, and Levskaya]{kitaev2020reformer}
N.~Kitaev, {\L}.~Kaiser, and A.~Levskaya.
\newblock Reformer: The efficient transformer.
\newblock \emph{arXiv preprint arXiv:2001.04451}, 2020.

\bibitem[Li et~al.(2024)Li, Yu, Li, and Zhu]{li2024functional}
T.~Li, B.~Yu, J.~Li, and Z.~Zhu.
\newblock Functional relation field: A model-agnostic framework for multivariate time series forecasting.
\newblock \emph{Artificial Intelligence}, 334:\penalty0 104158, 2024.

\bibitem[Li and Law(2024)]{li2024deep}
W.~Li and K.~E. Law.
\newblock Deep learning models for time series forecasting: a review.
\newblock \emph{IEEE Access}, 2024.

\bibitem[Liang et~al.(2024)Liang, Wen, Nie, Jiang, Jin, Song, Pan, and Wen]{liang2024foundation}
Y.~Liang, H.~Wen, Y.~Nie, Y.~Jiang, M.~Jin, D.~Song, S.~Pan, and Q.~Wen.
\newblock Foundation models for time series analysis: A tutorial and survey.
\newblock In \emph{Proceedings of the 30th ACM SIGKDD conference on knowledge discovery and data mining}, pages 6555--6565, 2024.

\bibitem[Liu et~al.(2024{\natexlab{a}})Liu, Guo, Dai, Li, Bao, Ren, Jiang, and Xia]{liu2024calf}
P.~Liu, H.~Guo, T.~Dai, N.~Li, J.~Bao, X.~Ren, Y.~Jiang, and S.-T. Xia.
\newblock Calf: Aligning llms for time series forecasting via cross-modal fine-tuning.
\newblock \emph{arXiv preprint arXiv:2403.07300}, 2024{\natexlab{a}}.

\bibitem[Liu et~al.(2022)Liu, Wu, Wang, and Long]{liu2022non}
Y.~Liu, H.~Wu, J.~Wang, and M.~Long.
\newblock Non-stationary transformers: Exploring the stationarity in time series forecasting.
\newblock \emph{Advances in Neural Information Processing Systems}, 35:\penalty0 9881--9893, 2022.

\bibitem[Liu et~al.(2023)Liu, Hu, Zhang, Wu, Wang, Ma, and Long]{liu2023itransformer}
Y.~Liu, T.~Hu, H.~Zhang, H.~Wu, S.~Wang, L.~Ma, and M.~Long.
\newblock itransformer: Inverted transformers are effective for time series forecasting.
\newblock \emph{arXiv preprint arXiv:2310.06625}, 2023.

\bibitem[Liu et~al.(2024{\natexlab{b}})Liu, Qin, Huang, Wang, and Long]{liu2024timerxl}
Y.~Liu, G.~Qin, X.~Huang, J.~Wang, and M.~Long.
\newblock Timer-xl: Long-context transformers for unified time series forecasting.
\newblock \emph{arXiv preprint arXiv:2410.04803}, 2024{\natexlab{b}}.

\bibitem[Liu et~al.(2024{\natexlab{c}})Liu, Zhang, Li, Huang, Wang, and Long]{liu2024timer}
Y.~Liu, H.~Zhang, C.~Li, X.~Huang, J.~Wang, and M.~Long.
\newblock Timer: Generative pre-trained transformers are large time series models.
\newblock \emph{arXiv preprint arXiv:2402.02368}, 2024{\natexlab{c}}.

\bibitem[Liu et~al.(2025)Liu, Qin, Huang, Wang, and Long]{liu2025autotimes}
Y.~Liu, G.~Qin, X.~Huang, J.~Wang, and M.~Long.
\newblock Autotimes: Autoregressive time series forecasters via large language models.
\newblock \emph{Advances in Neural Information Processing Systems}, 37:\penalty0 122154--122184, 2025.

\bibitem[Liu et~al.(2021)Liu, Zhu, Gao, and Xu]{liu2021forecast}
Z.~Liu, Z.~Zhu, J.~Gao, and C.~Xu.
\newblock Forecast methods for time series data: A survey.
\newblock \emph{Ieee Access}, 9:\penalty0 91896--91912, 2021.

\bibitem[Martin and Mahoney(2021)]{martin2021implicit}
C.~H. Martin and M.~W. Mahoney.
\newblock Implicit self-regularization in deep neural networks: Evidence from random matrix theory and implications for learning.
\newblock \emph{Journal of Machine Learning Research}, 22\penalty0 (165):\penalty0 1--73, 2021.

\bibitem[Martin et~al.(2021)Martin, Peng, and Mahoney]{martin2021predicting}
C.~H. Martin, T.~Peng, and M.~W. Mahoney.
\newblock Predicting trends in the quality of state-of-the-art neural networks without access to training or testing data.
\newblock \emph{Nature Communications}, 12\penalty0 (1):\penalty0 4122, 2021.

\bibitem[Montgomery et~al.(2015)Montgomery, Jennings, and Kulahci]{montgomery2015introduction}
D.~C. Montgomery, C.~L. Jennings, and M.~Kulahci.
\newblock \emph{Introduction to time series analysis and forecasting}.
\newblock John Wiley \& Sons, 2015.

\bibitem[{National Supercomputer Centre (NSC)}(2025)]{nsc_berzelius}
{National Supercomputer Centre (NSC)}.
\newblock Berzelius supercomputer, 2025.
\newblock URL \url{https://www.nsc.liu.se/systems/berzelius/}.
\newblock Accessed: 2025-03-04.

\bibitem[Nie et~al.(2022)Nie, Nguyen, Sinthong, and Kalagnanam]{nie2022time}
Y.~Nie, N.~H. Nguyen, P.~Sinthong, and J.~Kalagnanam.
\newblock A time series is worth 64 words: Long-term forecasting with transformers.
\newblock \emph{arXiv preprint arXiv:2211.14730}, 2022.

\bibitem[Shi et~al.(2025)Shi, Wang, Chen, Hu, Peng, and Shi]{shi2025digital}
S.~Shi, N.~Wang, S.~Chen, B.~Hu, J.~Peng, and Z.~Shi.
\newblock Digital mapping of soil salinity with time-windows features optimization and ensemble learning model.
\newblock \emph{Ecological Informatics}, 85:\penalty0 102982, 2025.

\bibitem[Shi et~al.(2024)Shi, Wang, Nie, Li, Ye, Wen, and Jin]{shi2024time}
X.~Shi, S.~Wang, Y.~Nie, D.~Li, Z.~Ye, Q.~Wen, and M.~Jin.
\newblock Time-moe: Billion-scale time series foundation models with mixture of experts.
\newblock \emph{arXiv preprint arXiv:2409.16040}, 2024.

\bibitem[Touvron et~al.(2023)Touvron, Lavril, Izacard, Martinet, Lachaux, Lacroix, Rozi{\`e}re, Goyal, Hambro, Azhar, et~al.]{touvron2023llama}
H.~Touvron, T.~Lavril, G.~Izacard, X.~Martinet, M.-A. Lachaux, T.~Lacroix, B.~Rozi{\`e}re, N.~Goyal, E.~Hambro, F.~Azhar, et~al.
\newblock Llama: Open and efficient foundation language models.
\newblock \emph{arXiv preprint arXiv:2302.13971}, 2023.

\bibitem[UVA-MLSys(2024)]{sa_timeseries}
UVA-MLSys.
\newblock Sa-timeseries: Self-attention time series models, 2024.
\newblock URL \url{https://github.com/UVA-MLSys/SA-Timeseries}.
\newblock Last accessed: 15 Nov 2024.

\bibitem[Vaswani(2017)]{vaswani2017attention}
A.~Vaswani.
\newblock Attention is all you need.
\newblock \emph{Advances in Neural Information Processing Systems}, 2017.

\bibitem[Wen et~al.(2022)Wen, Zhou, Zhang, Chen, Ma, Yan, and Sun]{wen2022transformers}
Q.~Wen, T.~Zhou, C.~Zhang, W.~Chen, Z.~Ma, J.~Yan, and L.~Sun.
\newblock Transformers in time series: A survey.
\newblock \emph{arXiv preprint arXiv:2202.07125}, 2022.

\bibitem[Woo et~al.(2022)Woo, Liu, Sahoo, Kumar, and Hoi]{woo2022etsformer}
G.~Woo, C.~Liu, D.~Sahoo, A.~Kumar, and S.~Hoi.
\newblock Etsformer: Exponential smoothing transformers for time-series forecasting.
\newblock \emph{arXiv preprint arXiv:2202.01381}, 2022.

\bibitem[Woo et~al.(2024)Woo, Liu, Kumar, Xiong, Savarese, and Sahoo]{woo2024unified}
G.~Woo, C.~Liu, A.~Kumar, C.~Xiong, S.~Savarese, and D.~Sahoo.
\newblock Unified training of universal time series forecasting transformers.
\newblock In \emph{PMLR}, 2024.

\bibitem[Wu et~al.(2022)Wu, Lian, Zeng, Xu, and Su]{wu2022aggregated}
Y.~Wu, C.~Lian, Z.~Zeng, B.~Xu, and Y.~Su.
\newblock An aggregated convolutional transformer based on slices and channels for multivariate time series classification.
\newblock \emph{IEEE Transactions on Emerging Topics in Computational Intelligence}, 7\penalty0 (3):\penalty0 768--779, 2022.

\bibitem[Yang et~al.(2022)Yang, Hu, Babuschkin, Sidor, Liu, Farhi, Ryder, Pachocki, Chen, and Gao]{yang2022tensor}
G.~Yang, E.~J. Hu, I.~Babuschkin, S.~Sidor, X.~Liu, D.~Farhi, N.~Ryder, J.~Pachocki, W.~Chen, and J.~Gao.
\newblock Tensor programs v: Tuning large neural networks via zero-shot hyperparameter transfer.
\newblock \emph{arXiv preprint arXiv:2203.03466}, 2022.

\bibitem[Yang et~al.(2024)Yang, Zhou, Tang, Li, and Hu]{yang2024breaking}
Z.~Yang, B.~Zhou, X.~Tang, R.~Li, and S.~Hu.
\newblock Breaking the weak semantics bottleneck of transformers in time series forecasting.
\newblock In \emph{ECAI 2024}, pages 1430--1437, 2024.

\bibitem[Yaprakdal and Varol~Ar{\i}soy(2023)]{yaprakdal2023multivariate}
F.~Yaprakdal and M.~Varol~Ar{\i}soy.
\newblock A multivariate time series analysis of electrical load forecasting based on a hybrid feature selection approach and explainable deep learning.
\newblock \emph{Applied Sciences}, 13\penalty0 (23):\penalty0 12946, 2023.

\bibitem[Zeng et~al.(2023)Zeng, Chen, Zhang, and Xu]{zeng2023transformers}
A.~Zeng, M.~Chen, L.~Zhang, and Q.~Xu.
\newblock Are transformers effective for time series forecasting?
\newblock In \emph{Proceedings of the AAAI conference on artificial intelligence}, volume~37, pages 11121--11128, 2023.

\bibitem[Zerveas et~al.(2021)Zerveas, Jayaraman, Patel, Bhamidipaty, and Eickhoff]{zerveas2021transformer}
G.~Zerveas, S.~Jayaraman, D.~Patel, A.~Bhamidipaty, and C.~Eickhoff.
\newblock A transformer-based framework for multivariate time series representation learning.
\newblock In \emph{Proceedings of the 27th ACM SIGKDD conference on knowledge discovery \& data mining}, pages 2114--2124, 2021.

\bibitem[Zhang et~al.(2024{\natexlab{a}})Zhang, Liu, Bai, and Li]{zhang2024hybrid}
J.~Zhang, H.~Liu, W.~Bai, and X.~Li.
\newblock A hybrid approach of wavelet transform, arima and lstm model for the share price index futures forecasting.
\newblock \emph{The North American Journal of Economics and Finance}, 69:\penalty0 102022, 2024{\natexlab{a}}.

\bibitem[Zhang et~al.(2024{\natexlab{b}})Zhang, Wen, Zhang, Cai, Jin, Liu, Zhang, Liang, Pang, Song, et~al.]{zhang2024self}
K.~Zhang, Q.~Wen, C.~Zhang, R.~Cai, M.~Jin, Y.~Liu, J.~Y. Zhang, Y.~Liang, G.~Pang, D.~Song, et~al.
\newblock Self-supervised learning for time series analysis: Taxonomy, progress, and prospects.
\newblock \emph{IEEE Transactions on Pattern Analysis and Machine Intelligence}, 2024{\natexlab{b}}.

\bibitem[Zhang and Yan(2023)]{zhang2023crossformer}
Y.~Zhang and J.~Yan.
\newblock Crossformer: Transformer utilizing cross-dimension dependency for multivariate time series forecasting.
\newblock In \emph{The eleventh international conference on learning representations}, 2023.

\bibitem[Zhang et~al.(2024{\natexlab{c}})Zhang, Li, and Liu]{zhang2024multivariate}
Z.~Zhang, W.~Li, and H.~Liu.
\newblock Multivariate time series forecasting by graph attention networks with theoretical guarantees.
\newblock In \emph{International Conference on Artificial Intelligence and Statistics}, pages 2845--2853. PMLR, 2024{\natexlab{c}}.

\bibitem[Zhou et~al.(2022)Zhou, Ma, Wen, Wang, Sun, and Jin]{zhou2022fedformer}
T.~Zhou, Z.~Ma, Q.~Wen, X.~Wang, L.~Sun, and R.~Jin.
\newblock Fedformer: Frequency enhanced decomposed transformer for long-term series forecasting.
\newblock In \emph{International conference on machine learning}, pages 27268--27286. PMLR, 2022.

\bibitem[Zhou et~al.(2023)Zhou, Niu, Sun, Jin, et~al.]{zhou2023one}
T.~Zhou, P.~Niu, L.~Sun, R.~Jin, et~al.
\newblock One fits all: Power general time series analysis by pretrained lm.
\newblock \emph{Advances in neural information processing systems}, 36:\penalty0 43322--43355, 2023.

\end{thebibliography}
\end{document}